\documentclass[10pt,twocolumn,letterpaper]{article}
\usepackage[pagenumbers]{iccv} 
\definecolor{iccvblue}{rgb}{0.21,0.49,0.74}
\usepackage[pagebackref,breaklinks,colorlinks,allcolors=iccvblue]{hyperref}
\usepackage{multirow}
\usepackage{colortbl}%
\newcommand{\myrowcolour}{\rowcolor[gray]{0.925}}

\definecolor{LightGray}{gray}{0.925}
\newcommand{\secondnum}[1]{\textcolor{blue}{\underline{#1}}}
\newcommand{\bestnum}[1]{\textcolor{red}{#1}}
\usepackage{makecell}
\usepackage{bbding}
\usepackage{algorithmic}
\usepackage{algorithm}
\newcommand\ours{TP-Diff\xspace}

\title{Learning Deblurring Texture Prior from Unpaired Data with Diffusion Model}
\author{Chengxu Liu\textsuperscript{1}\quad\quad Lu Qi\textsuperscript{2,3}\quad\quad Jinshan Pan\textsuperscript{4}\quad\quad Xueming Qian\textsuperscript{1}\quad\quad Ming-Hsuan Yang\textsuperscript{5}\\
\textsuperscript{1}Xi’an Jiaotong University\quad 
\textsuperscript{2}Wuhan University\quad
\textsuperscript{3}Insta360 \\
\textsuperscript{4}Nanjing University of Science and Technology\quad
\textsuperscript{5}University of California, Merced 
\\
}

\begin{document}
\maketitle
\begin{abstract}
Since acquiring large amounts of realistic blurry-sharp image pairs is difficult and expensive, learning blind image deblurring from unpaired data is a more practical and promising solution. Unfortunately, dominant approaches rely heavily on adversarial learning to bridge the gap from blurry domains to sharp domains, ignoring the complex and unpredictable nature of real-world blur patterns. In this paper, we propose a novel diffusion model (DM)-based framework, dubbed \ours, for image deblurring by learning spatially varying texture prior from unpaired data. In particular, \ours performs DM to generate the prior knowledge that aids in recovering the textures of blurry images. To implement this, we propose a Texture Prior Encoder (TPE) that introduces a memory mechanism to represent the image textures and provides supervision for DM training. To fully exploit the generated texture priors, we present the Texture Transfer Transformer layer (TTformer), in which a novel Filter-Modulated Multi-head Self-Attention (FM-MSA) efficiently removes spatially varying blurring through adaptive filtering. Furthermore, we implement a wavelet-based adversarial loss to preserve high-frequency texture details. Extensive evaluations show that \ours provides a promising unsupervised deblurring solution and outperforms SOTA methods in widely-used benchmarks.

\end{abstract}

\section{Introduction}
\label{sec:intro}

\begin{figure}[t]
  \centering
  \includegraphics[width=1.0\linewidth,page=1]{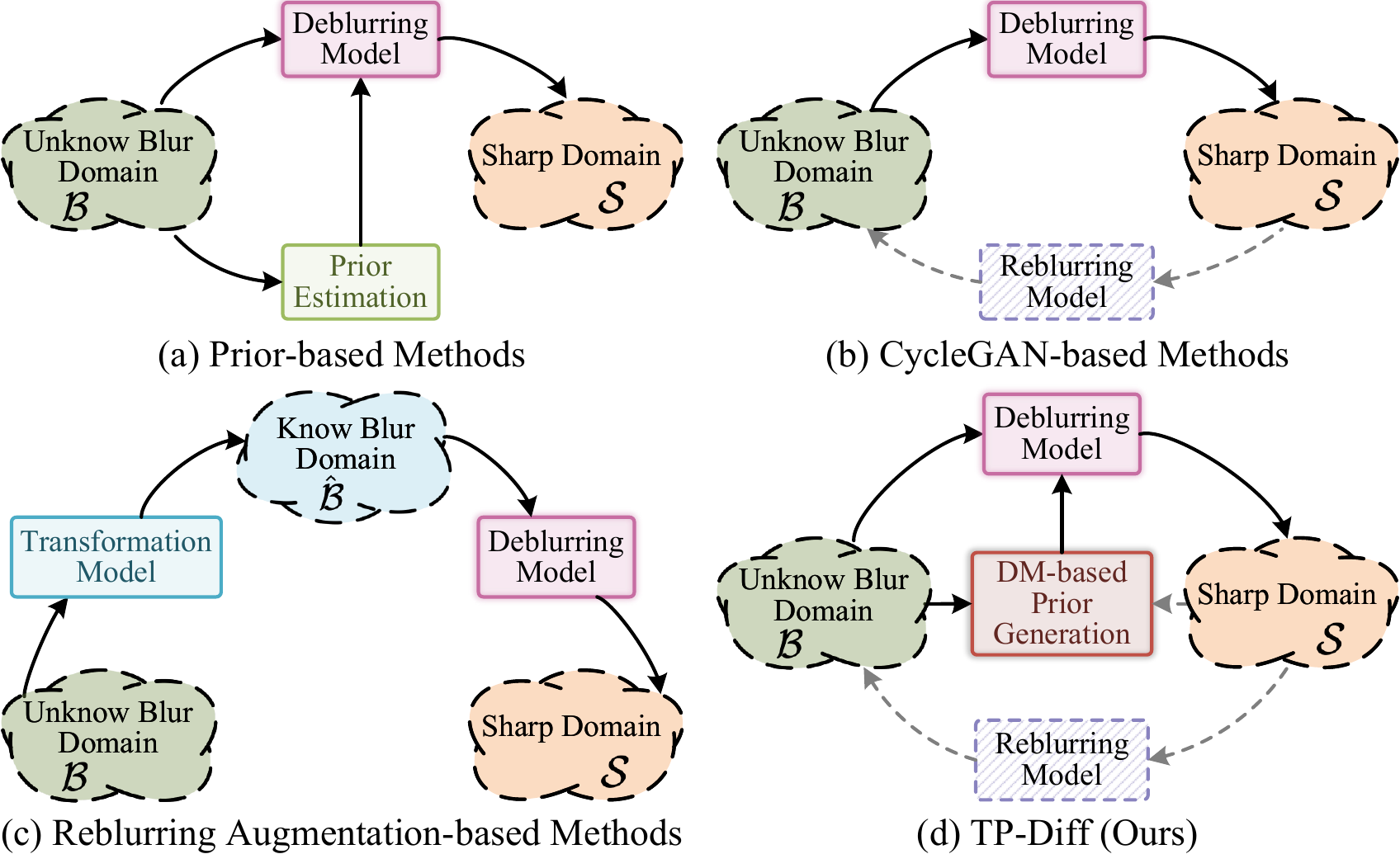}
  \vspace{-6mm}
   \caption{Illustration of previous unsupervised deblurring methods and our framework. (d) Our \ours leverages the DMs to denoise pure Gaussian noise into spatially varying texture prior for deblurring tasks. The dotted line indicates use only during training.}
   \label{fig:teaser}
   \vspace{-2mm}
\end{figure}

Image deblurring has been a topic of significant interest. The diversity and spatial variability of blurs pose significant challenges for developing effective solutions.
Early conventional approaches~\cite{chen2019blind,joshi2009image,pan2014deblurring,pan2016blind,yan2017image} aim to find the prior distribution of sharp images and maximize the posterior probability, yet lack generalization ability.

Recent years have witnessed a growing interest in deep learning. Most approaches~\cite{chen2022simple,kong2023efficient,liu2024motion,fang2023self} use synthetic paired data to supervise the DNN, achieving superior performance. However, the real-world blur is complex and unpredictable, making it infeasible to simulate real paired data manually. 
While recent studies~\cite{rim2020real,zhong2020efficient} have developed paired datasets using a dual-camera system, it is both expensive and time-consuming. In addition, it risks causing the model to overfit the specific characteristics of the camera used. Therefore, learning deblurring directly from unpaired blurry-sharp data presents a promising solution.

Existing unsupervised deblurring methods can be categorized into three paradigms.
1) Prior-based methods~\cite{dong2021learning,ren2020neural,zhang2023neural,tang2023uncertainty,jiang2023uncertainty} aim to train a sub-network that estimates prior knowledge for blur removal, as illustrated in Fig.~\ref{fig:teaser}(a). 
However, it is nearly impossible to find proper prior knowledge to model various blurs simultaneously.
2) Reblurring augmentation-based methods~\cite{pham2024blur2blur,wu2024IDBlau} employ a transformation model to convert blurs from an unknown domain to a known one, thus reducing the difficulty of the deblurring process, as illustrated in Fig.~\ref{fig:teaser}(c). Although favorable results are achieved, these methods are limited by the ability of existing deblurring models and incur additional computational costs.
3) CycleGAN-based methods~\cite{liu2017unsupervised,lu2019unsupervised,du2020learning,zhao2022fcl,chen2024unsupervised} construct blurry-sharp conversion cycles to learn the mapping between these two domains, as illustrated in Fig.~\ref{fig:teaser}(b). Unfortunately, these approaches overlook the spatial diversity of blur degrees and often tend to overfit a single blur template.
These limitations hinder the advancement and effectiveness of unsupervised image deblurring.

Recently, diffusion models (DMs) have shown impressive performance in image generation~\cite{ho2022video,rombach2022high,ho2020denoising,song2021denoising}. 
In image deblurring, most approaches use DMs directly for synthesizing sharp images, which eliminates blur but also introduces unpredictable artifacts~\cite{whang2022deblurring,ren2023multiscale,kawar2022denoising}.
In contrast, HiDiff~\cite{chen2024hierarchical} proposes using DMs to estimate a latent prior representation of sharp images for assisting the deblurring. 
Nevertheless, such prior is spatially out-of-order with a specific quantity and cannot be learned from unpaired data.
Therefore, this inspires us to address the problem of \textit{how to enable DMs to learn a spatially varying texture prior and exploit it in the unpaired deblurring task.}

To this end, we propose a novel diffusion model-based framework for unsupervised image deblurring (\ours). The key insight is enabling the DM to learn the spatially varying texture prior from unpaired data and assist the deblurring process, as illustrated in~\cref{fig:teaser}(d).
In particular, we follow~\cite{zhao2022fcl,chen2024unsupervised} to construct the cycle structure between sharp and blurry domains to learn the mapping relationships.
The training process of \ours involves two stages, as illustrated in~\cref{fig:overview}(b). In the first stage, we use the proposed Texture Prior Encoder (TPE) to learn texture representations from extensive unpaired data and encode them to obtain texture priors in latent space. 
In addition, we propose a Texture Transfer Transformer (TTformer) layer within the deblurring network to employ the learned texture priors effectively.
In the second stage, we freeze the parameters of TPE and train the DM to generate the texture prior using the TPE's output as ground truth.
During inference, the denoising network produces the texture prior from pure Gaussian noise, which guides the deblurring network. 
Compared to generating sharp images directly, generating the texture prior in latent space requires fewer iterations~\cite{chen2024hierarchical,xia2023diffir}.

Our contributions are summarized as follows:
\begin{itemize}
    \item We propose \ours for unsupervised image deblurring, in which the DM can learn spatially varying texture prior efficiently for blur removal. To our knowledge, this is the first work to integrate DM into unpaired restoration tasks.
    \item We propose a TPE that introduces a memory mechanism to encode texture priors of blurry images in latent space.
    \item We propose a TTformer layer, where a Filter-Modulated Multi-head Self-Attention (FM-MSA) employs adaptive filtering to remove spatially varying blurs.
    \item We propose a wavelet-based adversarial loss that helps preserve high-frequency texture details during training.
    \item Extensive experiments demonstrate that \ours outperforms SOTA methods on widely-used benchmarks.
\end{itemize}

\section{Related Work}
\label{sec:relat}

\subsection{Deep Supervised Image Deblurring}
Benefiting from the synthetic large-scale paired data, deep learning has achieved significant success in supervised image deblurring. 
Among them, some prior-free methods~\cite{zamir2021multi,cho2021rethinking,tsai2022stripformer,zamir2022restormer,li2022learning,chen2022simple,kong2023efficient,mao2023intriguing} attempt to develop more robust models that directly learn to remove blur using multi-scale learning and supervision~\cite{zamir2021multi,cho2021rethinking,chen2022simple}, attention mechanism~\cite{zamir2022restormer,tsai2022stripformer}, and frequency domain learning~\cite{kong2023efficient,mao2023intriguing}. 
However, the shift-invariant of CNNs limits their effectiveness in handling spatially varying blurs. 

To tackle this issue, prior-related methods~\cite{chen2024hierarchical,fang2023self,liu2024motion,kim2024real} learn the blur prior to guiding the deblurring network. Typically, UFPNet~\cite{fang2023self} and HiDiff~\cite{chen2024hierarchical} predict the prior representations through flow-based and diffusion-based models, respectively.
Although achieving significant performance, they are infeasible to learn the prior from unpaired data. 
Significantly, such practice of predicting the prior from the inputs is treated as an inverse problem of blur patterns, providing inspiration for addressing spatially varying blurs.

\begin{figure*}[th]
  \centering
    \includegraphics[width=1.0\linewidth,page=2]{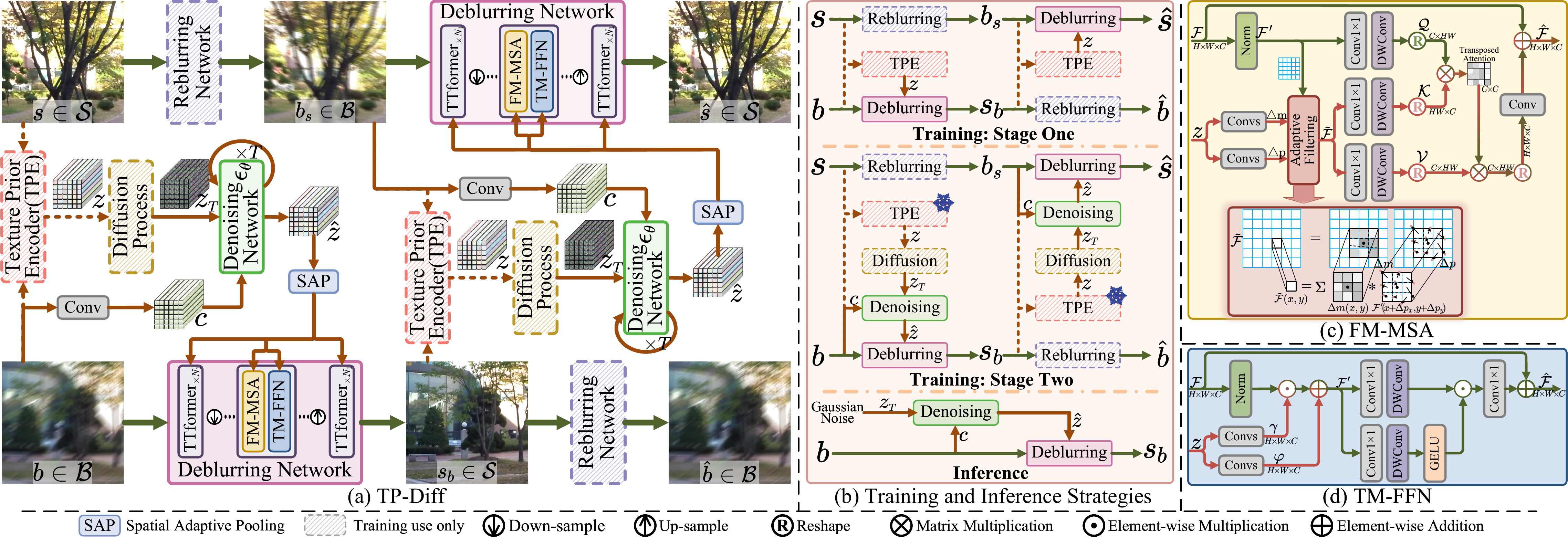}
    \vspace{-6mm}
   \caption{(a) Overview of \ours. The Texture Prior Encoder (TPE) is used for extracting texture prior from unpaired data. The diffusion process and denoising network are used for adding noise and generating the texture prior, respectively. (b) Schematic of training and inference strategies. (c) Filter-Modulated Multi-head Self-Attention (FM-MSA) and (d) Transform-Modulated Feed-Forward Network (TM-FFN) form the Texture Transfer Transformer layer (TTformer), which is used to exploit the texture prior in the deblurring network.}
   \label{fig:overview}
   \vspace{-2mm}
\end{figure*}

\subsection{Deep Unsupervised Image Deblurring}
Unsupervised image deblurring is directly trained using unpaired data. 
The absence of pixel-level constraints makes it challenging. 
Existing approaches can be categorized into three paradigms: prior-based, reblurring augmentation-based, and CycleGAN-based methods. 
According to the distribution characteristics of sharp data, the prior-based methods~\cite{dong2021learning,ren2020neural,zhang2023neural} use maximum the posterior probability to estimate the prior for the deblurring process. 
However, it is almost impossible to deal with multiple blurs simultaneously by assuming prior knowledge.
To avoid training unpaired data directly, reblurring augmentation-based methods~\cite{pham2024blur2blur,wu2024IDBlau,kimcontrollable} train an additional transformation model to convert the input from an unknown blur domain to a known blur domain~\cite{pham2024blur2blur} or generate more diverse blur data~\cite{wu2024IDBlau}. Nevertheless, these methods still rely on the pre-trained deblurring models that were trained on synthetic paired data.

To directly implement training based on unpaired data, CycleGAN-based methods~\cite{liu2017unsupervised,lu2019unsupervised,du2020learning,zhao2022fcl,chen2024unsupervised} construct sharp-blurry-sharp and blurry-sharp-blurry conversion cycles to learn the mapping between sharp and blurry domains by adversarial learning~\cite{goodfellow2014generative}. 
The recent SEMGUD~\cite{chen2024unsupervised} proposes a self-enhancement deblurring strategy to progressively improve the generated pseudo-paired data and reconstructor, achieving advanced results. 
However, this self-enhancement strategy depends on a fully supervised model trained on synthetic paired data and overlooks the spatial diversity of realistic blurs.
In our work, \ours enables tailored removal of various blurs without using any paired data by learning texture prior in different regions.

\subsection{Diffusion Model}
Recently, diffusion model (DM)~\cite{ho2020denoising,song2021denoising} has been attracting increasing attention in low-level vision tasks, such as super-resolution~\cite{ye2024learning,li2024rethinking}, denoising~\cite{kulikov2023sinddm}, inpainting~\cite{lugmayr2022repaint}, and so on.
In image deblurring, some works~\cite{whang2022deblurring,ren2023multiscale,kawar2022denoising} directly use the DM to generate sharp images from Gaussian noise through a stochastic iterative denoising process. Although this practice can generate clear textures, it also introduces unpredictable artifacts. Other works~\cite{xia2023diffir,chen2024hierarchical} use DM to estimate the latent prior representations of sharp images for assisting deblurring, yielding superior results while avoiding artifacts.
Nevertheless, these methods ignore the spatial diversity of blur in real-world scenarios, and it is also unavailable to train with unpaired images. In this paper, we apply DM to learn spatially varying texture prior from unpaired data for deblurring, which is the first work to use DM for unpaired image reconstruction.

\begin{figure*}[t]
  \centering
    \includegraphics[width=0.9\linewidth,page=3]{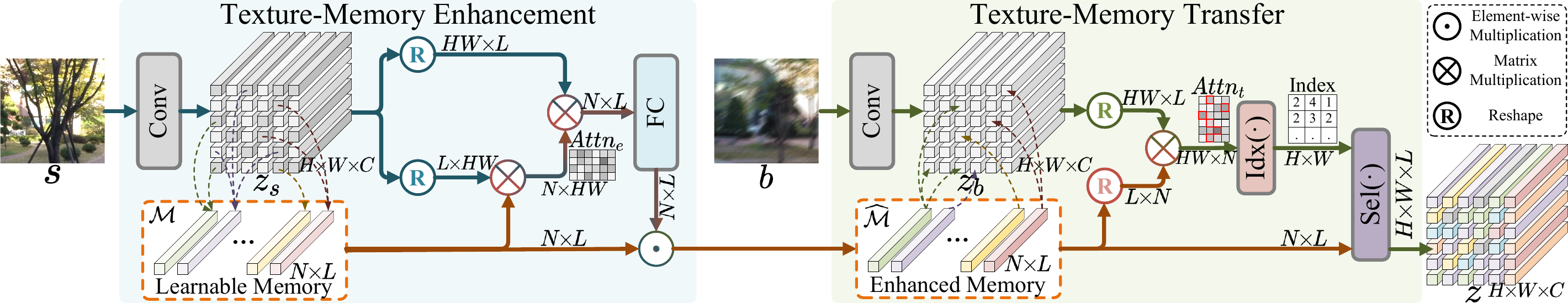}
    \vspace{-2mm}
   \caption{Structure of the Texture Prior Encoder (TPE), which serves to get spatially varying texture priors from the unpaired blurry-sharp input. $\mathrm{Idx}(\cdot)$ indicates that extracting the index with the highest score. $\mathrm{Sel}(\cdot)$ indicates that selecting the tokens according to the index.}
   \label{fig:tt}
   \vspace{-2mm}
\end{figure*}

\section{Methodology}
\label{sec:metho}

\subsection{Overall Architecture}
\label{overall}

As illustrated in~\cref{fig:overview}(a), our \ours mainly comprises four components: Texture Prior Encoder (TPE), Diffusion Model (DM), deblurring network, and reblurring network. Specifically, TPE is used to obtain the texture prior, which serves as the ground truth for DM training. DM consists of a diffusion process and a denoising network for adding noise and generating texture prior, respectively.
The deblurring network and reblurring network together form the entire cycle structure designed for removing and synthesizing blur, respectively. Within the deblurring network, the proposed Texture Transfer Transformer (TTformer) layer leverages texture prior knowledge to effectively remove blur.

To train \ours effectively, we follow established practices~\cite{chen2024hierarchical,li2024rethinking} and divide the training process into two stages, as illustrated in~\cref{fig:overview}(b). 
Taking the blurry image $b\in \mathcal{B}$ and unpaired sharp image $s\in \mathcal{S}$ as inputs. In stage one, we utilize the TPE to obtain spatially varying texture prior $z\in \mathbb{R}^{H\times W\times C}$ in latent space, and jointly train it and the entire cycle structure for blur removal. At this stage, the $z$ output from TPE is directly fed into the deblurring network without involving the diffusion and denoising processes. In stage two, we freeze the parameters of TPE, and joint train DM and the entire cycle structure.
This enables the denoising network to generate a more reliable spatially varying texture prior $\hat{z}\in \mathbb{R}^{H\times W\times C}$. 
At this stage, the $z$ output from TPE first adds noise to output $z_T\in \mathbb{R}^{H\times W\times C}$ through the diffusion process, and then generates the prior $\hat{z}$ through multiple iterations of the denoising process. $H\times W\times C$ is the size of the prior. Please refer to the supplementary for more detailed algorithms and structures. Next, we describe the training strategy and the inference process in order.

\subsection{Stage One: Latent Texture Prior Extraction}
In this stage, we aim to obtain a high-quality texture prior that guides the blur removal.
Specifically, we use TPE, which introduces a learnable memory bank, to learn texture representations from extensive unpaired data during training, thus extracting spatially varying texture prior $z$. 
Then, to fully leverage the texture prior and enhance the model capacity, we incorporate the TTformer at multiple scales within the deblurring network and fed $z$ into each scale through spatial adaptive pooling (SAP). Below, we introduce the core components TPE and TTformer.

\vspace{1mm}
\noindent\textbf{Texture Prior Encoder.}
As shown in~\cref{fig:tt}, TPE consists of a texture-memory enhancement part and a texture-memory transfer part. In the enhancement part, we initialize a group of learnable memories, and extract tokens from texture-rich sharp images as reference texture templates to enhance the memories. In the transfer part, we search for similar tokens from the enhanced memories and use them to represent texture templates in the input blurry image.

Specifically, in the enhancement part, we first compress sharp image $s$ into a higher dimensional space by a convolutional layer and obtain the texture-rich tokens $z_s\in \mathbb{R}^{H\times W\times C}$. 
Then, we use $z_s$ to enhance the introduced learnable memory $\mathcal{M}\in \mathbb{R}^{N\times L}$. $N$ is the memory size, which is set to 256 in our model. $L$ is the memory dimension, which is equal to $C$. In detail, we generate a corresponding attention map for the $\mathcal{M}$, formulated as:
\begin{equation}
    Attn_e = \mathrm{SoftMax}(\mathcal{M}\otimes \mathrm{R}(z_s)),
\end{equation}
where $Attn_e\in \mathbb{R}^{N\times HW}$ is the similarity of each memory with the texture-rich tokens. $\mathrm{SoftMax}(\cdot)$, $\mathrm{R}(\cdot)$, and $\otimes$ are the softmax function, reshape operation, and matrix multiplication, respectively. Next, we can compute enhanced memories $\widehat{\mathcal{M}}$ from all texture-rich tokens $z_s$, formulated as:
\begin{equation}
    \widehat{\mathcal{M}} = \mathcal{M}\odot \mathrm{FC}(Attn_e\otimes \mathrm{R}(z_s)),
\end{equation}
where $\odot$ denotes the element-wise multiplication. $\mathrm{FC}(\cdot)$ is the full connection layer used to refine the tokens after attention aggregation. This design allows memory to learn the valuable texture templates in all sharp data.

In the transfer part, we also compress blurry image $b$ and obtain the blurry tokens $z_b\in \mathbb{R}^{H\times W\times C}$.
Then, we search for similar tokens from the enhanced memories that have similar texture templates to those in the blurry input. The generated attention map $Attn_t\in \mathbb{R}^{HW\times N}$ is given by:
\begin{equation}
    Attn_t = \mathrm{R}(z_b)\otimes \mathrm{R}(\widehat{\mathcal{M}}).
\end{equation}
We select the enhanced memory with the highest attention map score for each blurry token and then aggregate them into the TPE's output $z$, formulated as:
\begin{equation}
    z = \mathrm{R}(\mathrm{Sel}(\mathrm{Idx}(Attn_t);\widehat{\mathcal{M}}))
\end{equation}
where $\mathrm{Idx}(\cdot)$ means extracting the index with the highest sorting score along the rows in the attention map. $\mathrm{Sel}(\cdot)$ means selecting the corresponding tokens from the enhanced memories for aggregation. 

Different from the vanilla self-attention that takes a weighted sum of tokens, we use only the most relevant tokens to ensure that the selected memories represent regions with similar texture templates. 
Compared to directly learning in image space, the proposed prior in the feature space is more robust and stable. Moreover, the TPE is robust enough to benefit from different unpaired sharp images. We provide detailed analysis in~\cref{abl}.

\vspace{1mm}
\noindent\textbf{TTformer.} 
Existing Transformer-based restoration methods~\cite{zamir2022restormer,kong2023efficient,tsai2022stripformer} achieve superior performance in supervised learning, but not explored in unsupervised feature learning.
Therefore, we introduce the TTformer layer to efficiently integrate the obtained prior knowledge into the unsupervised deblurring.
It consists of two components, the filter-modulated multi-head self-attention (FM-MSA) and the transform-modulated feed-forward network (TM-FFN).

Specifically, as shown in~\cref{fig:overview}(c), for the input feature $\mathcal{F}$ and input prior $z$, FM-MSA exploits an adaptive filtering operation to effectively remove spatially varying blurs. In detail, we use two three-layer convolution $\mathrm{Convs}(\cdot)$ to predict the vertical and horizontal offsets $\Delta p\in \mathbb{R}^{H\times W\times 2K^2}$ and weight $\Delta m\in \mathbb{R}^{H\times W\times K^2}$ of the filter, formulated as:
\begin{equation}
    \Delta p, \Delta m =  \mathrm{Convs}(z), \mathrm{Convs}(z),
\end{equation}
where $K$ is the kernel size. Then, taking the pixel with coordinates $(x,y)$ in the filtered feature $\mathcal{\tilde{F}}\in \mathbb{R}^{H\times W\times C}$ as example, the filtering process is formulated as:
{\small \begin{equation}
    \mathcal{\tilde{F}}(x,y)\!=\!\!\! \sum_{\Delta p(x,y)}\!\!\!\!\Big(\Delta m(x,y) * \mathcal{F}'\big(x\!+\!\Delta p_x,y\!+\!\Delta p_y\big)\Big),
\end{equation}}
where $\mathcal{F}'=\mathrm{Norm}(\mathcal{F})$ represents the normalized input feature $\mathcal{F}\in \mathbb{R}^{H\times W\times C}$. Subsequently, the visual tokens can be expressed as follows:
\begin{equation}
\begin{aligned}
    \mathcal{Q}&=\mathrm{R}(\mathrm{PDConv}(\mathcal{F}')),\\
    \mathcal{K},\mathcal{V}=\mathrm{R}&(\mathrm{PDConv}(\mathcal{\tilde{F}})),\mathrm{R}(\mathrm{PDConv}(\mathcal{\tilde{F}})),
\end{aligned}
\end{equation}
where $\mathrm{PDConv}(\cdot)$ are the $1\times1$ point-wise convolution and $3\times3$ depth-wise convolution. Finally, we follow existing work~\cite{zamir2022restormer} to generate a transposed-attention map and perform dot-product with $\mathcal{V}$ to obtain the final output of FM-MSA $\mathcal{\hat{F}}$. This process can be described as:
\begin{equation}
    \mathcal{\hat{F}}= \mathrm{Conv}(\mathrm{R}(\mathrm{SoftMax}(\frac{\mathcal{Q}\otimes \mathcal{K}}{\sqrt{C}})\otimes\mathcal{V}))\oplus\mathcal{F},
\end{equation}
where $C$ is the dimension of tokens. $\oplus$ denotes the element-wise addition. 
The design of adaptive filtering can fully utilize the texture prior and significantly improve the model's ability to recover various textures.

In addition, as shown in~\cref{fig:overview}(d), we use TM-FFN, similar in~\cite{xia2023diffir,li2024rethinking}, to aggregate local features from the FM-MSA output. In detail, TM-FFN first uses texture prior $z$ to get dynamic parameters $\gamma,\varphi\in \mathbb{R}^{H\times W\times C}$, which are then used to modulate the input feature $\mathcal{F}$ by:
\begin{equation}
\begin{aligned}
    \mathcal{F}'&= \mathrm{Norm}(\mathcal{F})\odot\gamma\oplus\varphi,\\
    \gamma,\varphi&=\mathrm{Convs}(z),\mathrm{Convs}(z),
\end{aligned}
\end{equation}
where $\mathcal{F}'\in\mathbb{R}^{H\times W\times C}$ is the modulated feature. Finally, we adopt the gating mechanism to enhance feature encoding and get the output $\hat{\mathcal{F}}\in\mathbb{R}^{H\times W\times C}$ of TM-FFN by:
{\small \begin{equation}
    \hat{\mathcal{F}}=\mathrm{Conv}(\mathrm{GELU}(\mathrm{PDConv}(\mathcal{F}'))\odot\mathrm{PDConv}(\mathcal{F}'))\oplus\mathcal{F}.
\end{equation}}

\vspace{1mm}
\noindent\textbf{Optimization Objective.}
We aim to get the spatially varying texture prior $z$ by TPE and jointly train it with the entire cycle structure. To achieve this, we follow existing works~\cite{chen2024unsupervised,zhao2022fcl}, using the same adversarial loss $\mathcal{L}_{GAN}$ and cycle consistency loss $\mathcal{L}_{CYC}$ to supervise training. In addition, to preserve high-frequency texture detail as possible, we propose a wavelet-based adversarial loss, formulated as:
\begin{equation}
\begin{aligned}
        \mathcal{L}_{Wave} &= \mathbb{E}_{s\sim p_{\Phi(s)}}[\mathrm{log}D_{S}(\Phi(s))]\\
        &+\mathbb{E}_{b\sim p_{b}}[\mathrm{log}(1-D_{S}(\Phi(DN(b)))],
\label{losswave}
\end{aligned}
\end{equation}
where $\Phi(\cdot)$ denotes the extraction of high-frequency components using the wavelet transform. $DN(\cdot)$ is the deblurring network. $D_{S}(\cdot)$  is the discriminator and tries to maximize the distinction between deblurred images and sharp images. The full objective function for stage one $\mathcal{L}_{s1}$ is a weighted sum of the above losses, formulated as:
{\small \begin{equation}
\mathcal{L}_{s1}=\lambda_{GAN}\mathcal{L}_{GAN}+\lambda_{CYC}\mathcal{L}_{CYC}+\lambda_{Wave}\mathcal{L}_{Wave},
\label{equ:s1}
\end{equation}}
where the hyper-parameters $\lambda_{GAN}$, $\lambda_{CYC}$, and $\lambda_{Wave}$ control the importance of each term.

\begin{table*}[t]\small
\centering
\begin{tabular}{clcccccccccc}
\toprule
\multicolumn{2}{c}{\multirow{2}{*}{Methods}}       & \multicolumn{2}{c}{{GoPro}~\cite{nah2017deep}} & \multicolumn{2}{c}{{HIDE}~\cite{shen2019human}} & \multicolumn{2}{c}{{RealBlur-R}~\cite{rim2020real}} & \multicolumn{2}{c}{{RealBlur-J}~\cite{rim2020real}}   &  \multicolumn{2}{c}{{Overhead}} \\
\cmidrule[0.1pt](lr{0.125em}){3-4}\cmidrule[0.1pt](lr{0.125em}){5-6}\cmidrule[0.1pt](lr{0.125em}){7-8}\cmidrule[0.1pt](lr{0.125em}){9-10}\cmidrule[0.1pt](lr{0.125em}){11-12}
 &  &  \cellcolor{LightGray}PSNR & \cellcolor{LightGray}SSIM & \cellcolor{LightGray}PSNR & \cellcolor{LightGray}SSIM    & \cellcolor{LightGray}PSNR & \cellcolor{LightGray}SSIM & \cellcolor{LightGray}PSNR & \cellcolor{LightGray}SSIM  & \cellcolor{LightGray}\#Param(M) & \cellcolor{LightGray}Latency(s) \\
\midrule
\multirow{5}{*}{\rotatebox{90}{\makecell[c]{Supervised\\Training}}}
 &   MSDI-Net~\cite{li2022learning}    &    33.28& 0.964& 31.02& 0.940 &35.88& 0.952 &28.59& 0.869   & 135.4 & -   \\
 &   \cellcolor{LightGray}NAFNet~\cite{chen2022simple}    &    \cellcolor{LightGray}33.69 &\cellcolor{LightGray}0.967& \cellcolor{LightGray}31.32& \cellcolor{LightGray}0.943& \cellcolor{LightGray}35.50& \cellcolor{LightGray}0.953& \cellcolor{LightGray}28.32 &\cellcolor{LightGray}0.857  & \cellcolor{LightGray}67.9 & \cellcolor{LightGray}0.04   \\
 &   icDPM~\cite{ren2023multiscale}    &    33.20 &0.963& 30.96& 0.938& -&- &28.81& 0.872   & 52.0 & -   \\
 &   \cellcolor{LightGray}HI-Diff~\cite{chen2024hierarchical}    &  \cellcolor{LightGray}33.33 &\cellcolor{LightGray}0.964 &\cellcolor{LightGray}31.46 &\cellcolor{LightGray}0.945 &\cellcolor{LightGray}36.28 &\cellcolor{LightGray}0.958 &\cellcolor{LightGray}29.15& \cellcolor{LightGray}0.890& \cellcolor{LightGray}28.5 &  \cellcolor{LightGray}-  \\
 &   UFPNet~\cite{fang2023self}    &     34.06 &0.968& 31.74 &0.947& 36.25& 0.953 &29.87& 0.884   & 80.3 &  -  \\
\midrule
\multirow{12}{*}{\rotatebox{90}{\makecell[c]{Unpaired\\Training}}} &  \cellcolor{LightGray}CycleGAN~\cite{zhu2017unpaired}    &     \cellcolor{LightGray}22.54 &\cellcolor{LightGray}0.757 &\cellcolor{LightGray}21.81& \cellcolor{LightGray}0.675 &\cellcolor{LightGray}12.38& \cellcolor{LightGray}0.242 &\cellcolor{LightGray}19.79& \cellcolor{LightGray}0.633   & \cellcolor{LightGray}11.38 & \cellcolor{LightGray}0.02  \\
 &   UNIT~\cite{liu2017unsupervised}    &  22.58 & 0.778 & 22.21& 0.702& 28.94& 0.717 & 24.55& 0.755 & 11.38 & 0.03   \\ 
 &  \cellcolor{LightGray}UID-GAN~\cite{lu2019unsupervised}    &    \cellcolor{LightGray}23.56 &\cellcolor{LightGray}0.812 &\cellcolor{LightGray}22.70 &\cellcolor{LightGray}0.715 &\cellcolor{LightGray}16.64 &\cellcolor{LightGray}0.323 &\cellcolor{LightGray}22.87 &\cellcolor{LightGray}0.671  & \cellcolor{LightGray}19.93 &   \cellcolor{LightGray}0.04 \\
 &   LIR~\cite{du2020learning}    &    23.26& 0.818& 22.44& 0.708& 26.94& 0.644 & 23.59& 0.747  & 11.15 &  0.02  \\
 &   \cellcolor{LightGray}U-GAT-IT~\cite{kim2019u}    &  \cellcolor{LightGray}20.09& \cellcolor{LightGray}0.703& \cellcolor{LightGray}20.65& \cellcolor{LightGray}0.632& \cellcolor{LightGray}13.64& \cellcolor{LightGray}0.323 & \cellcolor{LightGray}16.84& \cellcolor{LightGray}0.604  & \cellcolor{LightGray}278.96 & \cellcolor{LightGray}-   \\ 
 &   DCD-GAN~\cite{chen2022unpaired}   &  21.99& 0.772& 23.19& 0.777& 17.27& 0.362 & 20.99& 0.671   & 11.38 &   0.02 \\
 &  \cellcolor{LightGray}FCL-GAN~\cite{zhao2022fcl}    &  \cellcolor{LightGray}24.59& \cellcolor{LightGray}0.831 &\cellcolor{LightGray}23.43 &\cellcolor{LightGray}0.782 &\cellcolor{LightGray}28.37& \cellcolor{LightGray}0.663& \cellcolor{LightGray}25.35 & \cellcolor{LightGray}0.736  & \cellcolor{LightGray}24.56 & \cellcolor{LightGray}0.01   \\
 &   UVCGANv2~\cite{torbunov2023uvcgan}    &  24.15& 0.814& 23.35& 0.792& 25.31& 0.658 & 24.82& 0.740  & 32.56 & 0.03   \\
 & \cellcolor{LightGray}UCL~\cite{wang2024ucl}    &  \cellcolor{LightGray}25.06& \cellcolor{LightGray}0.839& \cellcolor{LightGray}23.85& \cellcolor{LightGray}0.816& \cellcolor{LightGray}30.53& \cellcolor{LightGray}0.757 & \cellcolor{LightGray}26.04& \cellcolor{LightGray}0.784 & \cellcolor{LightGray}19.45 &  \cellcolor{LightGray}0.02  \\
 &   \ours    &    28.13   & 0.903   & 26.70  & 0.821    & 34.95 & 0.933  & \secondnum{28.01}& 0.836    & 11.89 &  0.04  \\
 \cmidrule(lr{0.125em}){2-12}
 &   \cellcolor{LightGray}SEMGUD~\cite{chen2024unsupervised}    &   \cellcolor{LightGray}\secondnum{29.06} &\cellcolor{LightGray}\secondnum{0.927} &\cellcolor{LightGray}\secondnum{27.64}& \cellcolor{LightGray}\secondnum{0.892}& \cellcolor{LightGray}\bestnum{35.51}& \cellcolor{LightGray}\secondnum{0.946}& \cellcolor{LightGray}\secondnum{28.01}& \cellcolor{LightGray}\bestnum{0.844}   & \cellcolor{LightGray}67.9 & \cellcolor{LightGray}0.04   \\
 &   \ours-\textit{se}    &    \bestnum{30.16}  & \bestnum{0.934}   & \bestnum{28.21} &  \bestnum{0.909}   & \secondnum{35.32} & \bestnum{0.947}  & \bestnum{28.03}&  \secondnum{0.843}   & 11.89 & 0.04   \\
\bottomrule
\end{tabular}
\vspace{-3mm}
\caption{Quantitative comparison on the GoPro~\cite{nah2017deep}, HIDE~\cite{shen2019human}, RealBlur-R~\cite{rim2020real}, and RealBlur-J~\cite{rim2020real}. 
Latency(s) is computed on images with the size of $256\times256$ with an NVIDIA RTX 3090 GPU. \textcolor{red}{Red} and \textcolor{blue}{\underline{blue}} indicate the best and second best performance, respectively.}
\vspace{-2mm}
\label{tab:main1}
\end{table*}

\subsection{Stage Two: Texture Prior Generation}
In this stage, we jointly train the DM and the entire cycle structure so that the denoising network generates effective texture priors to enhance the deblurring network. 
Specifically, following the existing works~\cite{chen2024hierarchical,li2024rethinking,xia2023diffir}, our DM consists of the forward diffusion process and the reverse denoising process.
In the forward diffusion process, we first adopt the TPE, in which the parameters are frozen, to generate the texture prior $z\in\mathbb{R}^{H\times W\times C}$ as ground truth, and then add noise to $z$ by the diffusion process and get the $z_T\in\mathbb{R}^{H\times W\times C}$ with the same distribution as the pure Gaussian noise.
In the reverse denoising process, we use the denoise network to generate the reconstructed texture prior $\hat{z}\in\mathbb{R}^{H\times W\times C}$ using the extracted feature $c\in\mathbb{R}^{H\times W\times C}$ by a convolutional layer from blurry input as the condition and $z$ as the target.

\vspace{1mm}
\noindent\textbf{Diffusion Process.}
Following existing works~\cite{ho2020denoising,song2021denoising}, we perform the forward Markov process starting from $z$, and gradually add Gaussian noise by $T$ iterations as follows:
\begin{equation}
    q(z_{T}\mid z)=\mathcal{N} (z_{T} ;\sqrt{\bar{\alpha}_{T}}z,(1-\bar{\alpha}_{T})\mathrm{I}),
\end{equation}
where $T$ is the total number of iteration steps. $\mathcal{N}(\cdot)$ denotes the Gaussian distribution. $\alpha=1-\beta_t$, $\bar{\alpha}_t=\prod_{i=1}^t \alpha_i$, where $t\in\{1,\dots,T\}$, $\beta_{1:T} \in (0,1)$, are hyper-parameters derived through iterative derivation with reparameterization~\cite{kingma2013auto} to control the amount of noise added at each step.

\vspace{1mm}
\noindent\textbf{Denoising Process.}
Aiming to generate the texture prior $\hat{z}$ from Gaussian noise, we perform the Markov chain that runs backward from $z_T$ to $z$, and gradually remove the noise by $T$ iterations.
In the inverse step from $z_t$ to $z_{t-1}$:
\begin{equation}
\begin{aligned}
q(z_{t-1} \mid z_t, z)=\mathcal{N}&(z_{t-1};\mu_t(z_t, z), \frac{1-\bar{\alpha}_{t-1}}{1-\bar{\alpha}_t} \beta_t \mathrm{I}),\\
\mu_t(z_t, z)=&\frac{1}{\sqrt{\alpha_t}}(z_t-\frac{1-\alpha_t}{\sqrt{1-\bar{\alpha}_t}} \epsilon),
\end{aligned}
\label{2}
\end{equation}
where $\epsilon$ represents the noise in $z_t$, which is the only uncertain variable that needs to be estimated at each step using the denoising network. Therefore, we use a neural network consisting of a series of stacked residual blocks, denoted as $\epsilon_\theta$, to estimate the noise with the condition $c$. Then, we further substitute $\epsilon_\theta$ into~\cref{2} to obtain:
\begin{equation}
z_{t-1}\!=\!\frac{1}{\sqrt{\alpha_t}}(z_t-\frac{1-\alpha_t}{\sqrt{1-\bar{\alpha}_t}} \epsilon_\theta(z_t, A_h, t))\!+\!\sqrt{1\!-\!\alpha_t} \epsilon_t,
\label{3}
\end{equation}
where $\epsilon_t\!\sim\!\mathcal{N}(0,\mathrm{I})$. $\epsilon_\theta(\mathrm{z}_t, c, t)$ is the noise estimated by the denoising network. By repeating the $T$ times sampling iterations in~\cref{3}, we can obtain the reconstructed texture prior $\hat{z}$. The purpose of using residual blocks as the denoising network is to ensure the same resolution of inputs and outputs while minimizing the model parameters. As shown in~\cref{fig:overview}(a), the output $\hat{z}$ from the denoising network is finally utilized to guide the deblurring network.
Our DM is more effective compared to other texture prior generation methods. We provide detailed analysis in~\cref{abl}.

\vspace{1mm}
\noindent\textbf{Optimization Objective.}
Our objective is to joint train the denoising network $\epsilon_\theta$ and the entire cycle structure. To achieve this, we additionally include the diffusion loss $\mathcal{L}_{diff}$ based on $\mathcal{L}_{s1}$ in~\cref{equ:s1}, formulated as:
\begin{equation}
\mathcal{L}_{s2}=\mathcal{L}_{s1} + \lambda_{diff}\mathcal{L}_{diff}, \ \   \mathcal{L}_{diff}=\left \| z-\hat{z} \right \| _1.
\label{loss2}
\end{equation}

\subsection{Inference}
During inference, only given a blurry input image $b$, we first extract the blur feature $c$ from the blurry input as the condition of denoising network. Then, we randomly sample pure Gaussian noise $z_T$. After $T$ times denoising process in~\cref{3}, denoising network generates the texture prior $\hat{z}$ using $z_T$ and $c$. Finally, we feed $\hat{z}$ to the deblurring network consisting of TTformer to compute the deblurred result.

\section{Experiments}
\label{sec:exper}
\subsection{Datasets and Metrics}
We evaluate the our method on widely-used datasets: {GoPro}~\cite{nah2017deep}, {HIDE}~\cite{shen2019human}, {RealBlur}~\cite{rim2020real}, {RB2V\!\_Street}~\cite{pham2023hypercut}, and {RSBlur}~\cite{rim2022realistic}. 
For fair comparisons, we follow existing works~\cite{chen2024unsupervised,pham2024blur2blur} to split the training set of GoPro, RB2V\!\_~\!Street, and RSBlur datasets into separate blurry and sharp image parts to constitute unpaired blurry-sharp pairs for training. We conduct three sets of experiments: i) Using the GoPro training set for training and the test sets for GoPro, HIDE, RealBlur-R, and RealBlur-J for testing. ii) Using the RB2V\!\_~\!Street training set for training and its test set for testing. iii) Using the RSBlur training set for training and its test set for testing.
We keep the same evaluation metrics of PSNR(dB) and SSIM as previous works~\cite{chen2024unsupervised,pham2024blur2blur}.

\begin{table}[t]\small
\setlength{\tabcolsep}{0.5mm}{
\centering
\begin{tabular}{lcc}
\toprule
Methods   & {RB2V\!\_~\!Street}~\cite{pham2023hypercut} & {RSBlur}~\cite{rim2022realistic}  \\
\midrule
\textit{Generalized Deblurring} & &\\
\myrowcolour
BSRGAN~\cite{zhang2021designing} &23.31/0.645 & 27.11/0.810  \\
RSBlur~\cite{rim2022realistic} &23.42/0.603 & 26.98/0.798  \\
\myrowcolour
NAFNet~\cite{chen2022simple} &28.72/0.883 & 33.06/0.888  \\
Restormer~\cite{zamir2022restormer} &27.43/0.849 & 32.87/0.874 \\
\midrule
\textit{Reblurring Augmentation}& &\\
\myrowcolour
NAFNet~\cite{chen2022simple}+Blur2Blur~\cite{pham2024blur2blur}&26.98/0.812 & 29.00/0.857 \\
Restormer~\cite{zamir2022restormer}+Blur2Blur~\cite{pham2024blur2blur}&25.97/0.750&28.89/0.850 \\
\midrule
\textit{Unpaired Training}& &\\
\myrowcolour
CycleGAN~\cite{zhu2017unpaired} &21.21/\secondnum{0.582} & 23.34/\bestnum{0.782}\\
DualGAN~\cite{yi2017DualGAN} & 21.02/0.556  &  22.78/0.704 \\
\myrowcolour
UNIT~\cite{liu2017unsupervised} &20.53/0.519 &26.49/0.734 \\
UID-GAN~\cite{lu2019unsupervised} & 22.03/0.564&26.48/0.713 \\
\myrowcolour
LIR~\cite{du2020learning} &20.43/0.534 & 24.44/0.720 \\
U-GAT-IT~\cite{kim2019u} & 20.75/0.539 & 22.17/0.629 \\
\myrowcolour
DCD-GAN~\cite{chen2022unpaired} & 21.20/0.537 & 25.90/0.704\\
FCL-GAN~\cite{zhao2022fcl} & 21.77/0.560& 28.17/0.746\\
\myrowcolour
UVCGANv2~\cite{torbunov2023uvcgan} & \secondnum{22.23}/0.561& 27.80/0.708\\
UCL~\cite{wang2024ucl} &21.38/0.556 & 26.81/0.717\\
\myrowcolour
\ours & \bestnum{22.89}/\bestnum{0.639} & \bestnum{28.40}/\secondnum{0.751}\\
\bottomrule
\end{tabular}
\vspace{-3mm}
\caption{Quantitative comparison (PSNR$\uparrow$ and SSIM$\uparrow$) on the RB2V\!\_~\!Street~\cite{pham2023hypercut} and RSBlur~\cite{rim2022realistic} datasets. 
}
\vspace{-3mm}
\label{tab:main2}}
\end{table}

\begin{figure*}[t]
  \centering
    \includegraphics[width=0.98\linewidth,page=4]{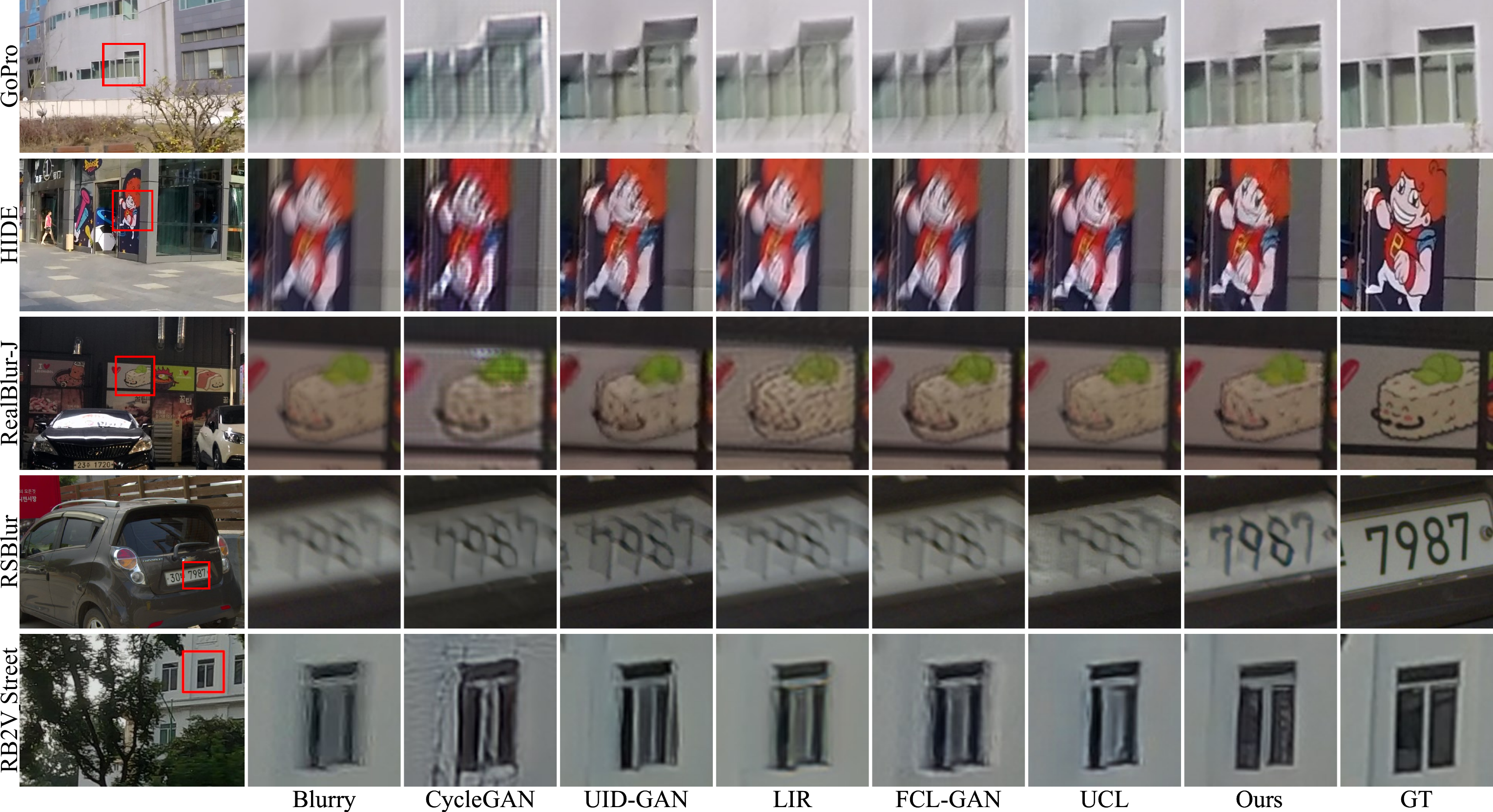}
    \vspace{-3mm}
   \caption{Visual results on {GoPro}~\cite{nah2017deep}, {HIDE}~\cite{shen2019human}, {Realblur-J}~\cite{rim2020real}, {RSBlur}~\cite{rim2022realistic}, and {RB2V\!\_~\!Street}~\cite{pham2023hypercut} datasets. The method is shown at the bottom of each case. Zoom in to see better visualization.}
   \label{fig:case}
\end{figure*}

\begin{figure*}[t]
  \begin{minipage}{0.75\textwidth}\small
    \centering
    \setlength{\tabcolsep}{1.3mm}{
    \begin{tabular}{l|cccccc|cc}
    \toprule
    \multirow{2}{*}{Methods}       & \multicolumn{6}{c|}{Components}  &   \multicolumn{2}{c}{Metrics} \\
    \cmidrule[0.1pt](lr{0.125em}){2-7}\cmidrule[0.1pt](lr{0.125em}){8-9}
     & \cellcolor{LightGray}Diffusion & \cellcolor{LightGray}TPE & \cellcolor{LightGray}TTformer  & \cellcolor{LightGray}Multi-Scale  &  \cellcolor{LightGray}Joint-Train & \cellcolor{LightGray}WaveLoss  & \cellcolor{LightGray}PSNR & \cellcolor{LightGray}SSIM   \\
    \midrule
    \textit{w/o} DM   &    \XSolidBrush  & \Checkmark & \Checkmark  & \Checkmark &  \XSolidBrush & \Checkmark  &26.46 & 0.867   \\
    \myrowcolour
    \textit{w/o} TPE   &  \Checkmark  & \XSolidBrush  & \Checkmark   & \Checkmark &  \Checkmark & \Checkmark  &27.36 & 0.886   \\
    \textit{w/o} TTformer   &    \Checkmark  & \Checkmark & \XSolidBrush   & \Checkmark &  \Checkmark & \Checkmark  &27.19 & 0.884   \\
    \myrowcolour
    \textit{w/o} MS   &    \Checkmark  & \Checkmark & \Checkmark    & \XSolidBrush &  \Checkmark & \Checkmark  &27.89 & 0.896   \\
    \textit{w/o} JT   &    \Checkmark  & \Checkmark & \Checkmark   & \Checkmark &  \XSolidBrush & \Checkmark  &27.65 & 0.896   \\
    \myrowcolour
    \textit{w/o} WaveLoss   &    \Checkmark  & \Checkmark & \Checkmark    & \Checkmark&  \Checkmark & \XSolidBrush  &28.01 & 0.899   \\
    Full model   &    \Checkmark  & \Checkmark & \Checkmark    & \Checkmark &  \Checkmark & \Checkmark  & \textbf{28.13} & \textbf{0.903}   \\
    \bottomrule
    \end{tabular}
    \vspace{-2mm}
    \captionof{table}{Ablation study of each component on the GoPro~\cite{nah2017deep} dataset.}
    \label{tab:ablation}}
  \end{minipage}%
  \hfill
  \begin{minipage}{0.25\textwidth}
  \centering
    \includegraphics[width=1.0\linewidth,page=8]{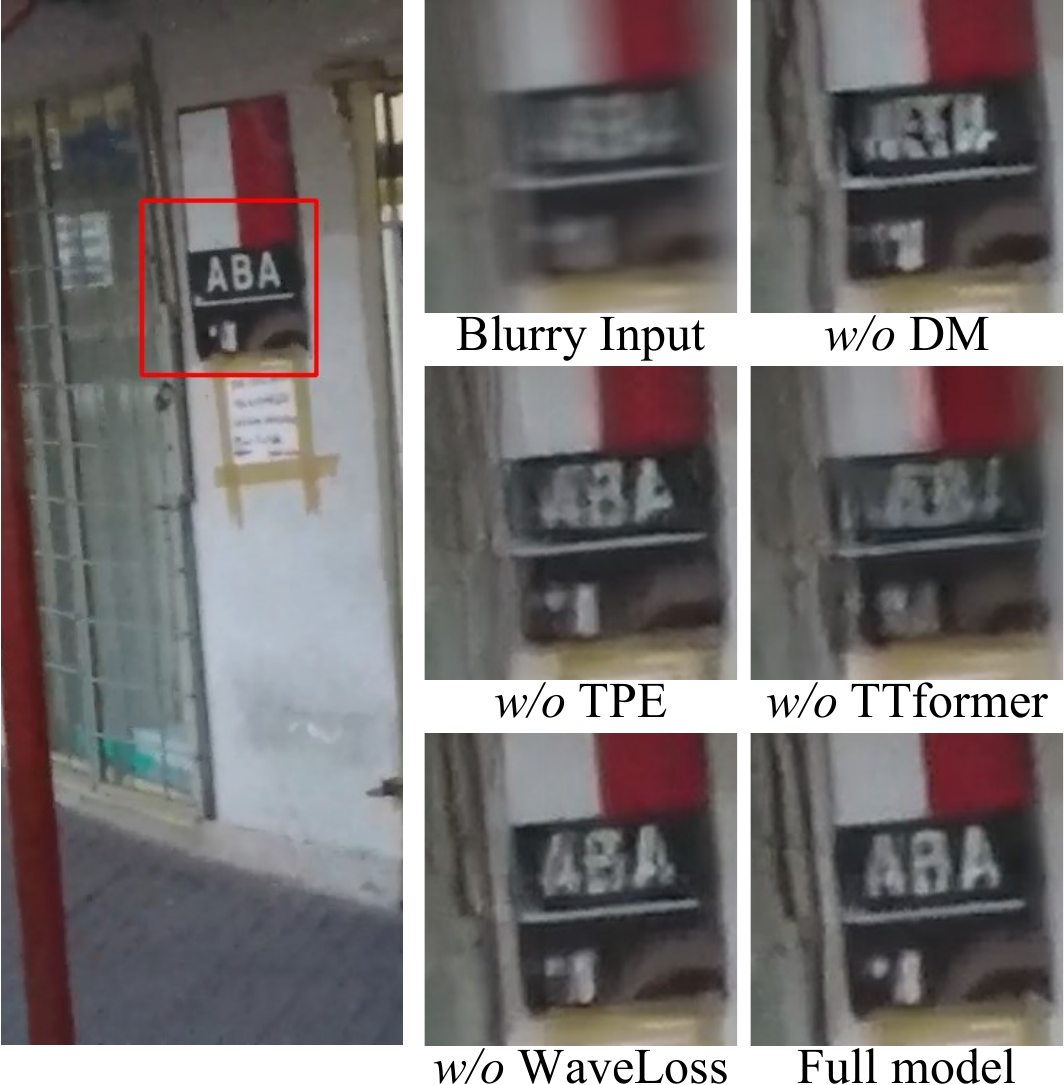}
    \vspace{-7mm}
   \caption{Visualization of ablation.}
   \label{fig:abl}
  \end{minipage}
\end{figure*}

\subsection{Implementation Details}
We follow existing works~\cite{zhu2017unpaired,chen2024unsupervised} to use the same discriminator and use the UNet~\cite{ronneberger2015u} as the reblurring network. In the deblurring network, we set the number of TTformer as [4,6,6,4], the attention heads as [1,2,4,8], and the number of channels $C$ as 48, and the kernel size $K$ of adaptive filtering as 5. In the denoising network, we set the number of ResBlock to 5 and the iteration step $T$ to 8. During training, the hyper-parameters $\lambda_{GAN}$, $\lambda_{CYC}$, $\lambda_{Wave}$, and $\lambda_{diff}$ are set to 1, 0.1, 0.2, and 1, respectively. We use Adam optimizer with $\beta_{1}=0.9$ and $\beta_{2}=0.999$, learning rate is $1\times 10^{-4}$ for both training stages. The epoch number for each stage is 200. We set the batch size as $8$ and the input patch size as $256\times 256$ and augment the data with random horizontal and vertical flips. All experiments were based on PyTorch and trained on an NVIDIA RTX A6000 GPU.

\subsection{Comparisons with State-of-the-art Methods}
We evaluate our method against 16 SOTA models~\cite{li2022learning,chen2022simple,ren2023multiscale,chen2024hierarchical,fang2023self,pham2024blur2blur,zhu2017unpaired,liu2017unsupervised,lu2019unsupervised,du2020learning,kim2019u,chen2022unpaired,zhao2022fcl,torbunov2023uvcgan,wang2024ucl,chen2024unsupervised}. 
For fair comparisons, we obtain the performance from their original paper or reproduce results by officially released codes.

\vspace{1mm}
\noindent\textbf{Quantitative Comparison.}
The performance comparisons on the GoPro~\cite{nah2017deep}, HIDE~\cite{shen2019human}, RealBlur-R~\cite{rim2020real}, and RealBlur-J~\cite{rim2020real} test sets are shown in~\cref{tab:main1}. \ours performs favorably against the latest unpaired training methods (\eg,~UCL~\cite{wang2024ucl}).
This is because our texture prior generated by DM can effectively handle spatially varying blurs.
Notably, although the latest SEMGUD~\cite{chen2024unsupervised} proposes a self-enhancement strategy that obtains favorable performance, this approach lacks fairness by introducing pre-trained fully supervised models to guide model training.
Therefore, we train another version of our model which is optimized with a similar strategy named \ours-\textit{se} for fair comparisons.
\cref{tab:main1} shows that our method also achieves better results while using fewer parameters proving its effectiveness.

In addition, we also verify the generalization capabilities of \ours on both realistic RB2V\!\_~\!Street~\cite{pham2023hypercut} and RSBlur~\cite{rim2022realistic} datasets. \cref{tab:main2} shows that \ours outperforms other unpaired training-based methods by 0.66 dB and 0.23 dB on RB2V\!\_~\!Street and RSBlur, respectively. The results verify that \ours has strong generalization capabilities and can effectively handle various realistic blurs through the generated texture prior.
Please refer to the supplementary for more quantitative results and efficiency analysis.

\vspace{1mm}
\noindent\textbf{Qualitative Comparison.}
We evaluate the visual quality of different methods in~\cref{fig:case}. The results show that \ours has a great improvement in visual quality, especially for areas with detailed textures. For example, in the first and fifth rows of~\cref{fig:case}, \ours can recover clearer windows. The results verify that \ours can exploit DM to generate texture priors that assist in producing finer results. Please refer to the supplementary for more visual results.

\begin{table}[t]\small
\setlength{\tabcolsep}{1.2mm}{
\centering
\begin{tabular}{l|l|c|c|cc}
\toprule
Exp. & Methods   &  Sampling &  $N$ & PSNR  & SSIM   \\
\midrule
\myrowcolour%
(a) & Latent encoder~\cite{chen2024hierarchical,li2024rethinking}& Random & - & 27.74 & 0.895 \\
\midrule
(b) &Enhancement part& Random& 256 & 27.73 & 0.895 \\
\myrowcolour%
(c) &Transfer part& Random& 256 & 27.88 & 0.897 \\
\midrule
(d) &TPE (RGB Space)& Random& 256 & 26.95 & 0.869 \\
\midrule
\cellcolor{LightGray}(e) &\multirow{5}{*}{TPE (Ours)} & \cellcolor{LightGray}Cluster  & \cellcolor{LightGray}256 & \cellcolor{LightGray}28.18 &  \cellcolor{LightGray}0.902 \\
(f) & & Singular & 256 & 27.94 & 0.898  \\
\cellcolor{LightGray}(g) & & \cellcolor{LightGray}{Random} & \cellcolor{LightGray}128 & \cellcolor{LightGray}27.99 & \cellcolor{LightGray}0.899 \\
(h) & & Random &256 & 28.13 &0.903 \\
\cellcolor{LightGray}(i) & &\cellcolor{LightGray}Random &\cellcolor{LightGray}512 &  \cellcolor{LightGray}28.19 &  \cellcolor{LightGray}0.904 \\
\bottomrule
\end{tabular}
\vspace{-2mm}
\caption{Ablation study of Texture Prior Encoder (TPE).}
\label{tab:te}}
\vspace{-2mm}
\end{table}

\subsection{Ablation Study}
\label{abl}
In this section, we explore the effectiveness of each key component. All ablation studies are trained in the same settings and evaluated on GoPro~\cite{nah2017deep} for fair comparisons.

\vspace{1mm}
\noindent\textbf{Effectiveness of Individual Components.}
We construct ablations in~\cref{tab:ablation} and~\cref{fig:abl} to demonstrate the effectiveness of each component in \ours.
We remove each component from the full model to compare performance and keep the same model size for fairness.
When DM is not used (``\textit{w/o} DM''), the PSNR is reduced by 1.67dB, which indicates that DM can generate valuable texture prior to enhance deblurring performance.
When we remove the TPE (``\textit{w/o} TPE'') and TTformer (``\textit{w/o} TTformer''), the PSNR decreases by 0.77dB and 0.94dB, respectively, and the visual quality is severely degraded. It proves that TPE can produce accurate texture prior, and TTformer can effectively remove blurs by exploiting the texture prior.
Performance also degrades when we remove the multi-scale learning (``\textit{w/o} MS), only training the DM in the second stage (``\textit{w/o} JT''), and remove the wavelet-based adversarial loss (``\textit{w/o} WaveLoss''). 
It demonstrates that these components are effective in improving model capability.
Overall, our full model achieves a PSNR of 28.13 dB with better visual quality, proving the effectiveness of each component.

\vspace{1mm}
\noindent\textbf{Effectiveness of TPE.}
We constructed several experiments in~\cref{tab:te} to demonstrate the effectiveness of TPEs.
(a) and (h) show that the TPE is more conducive to extracting texture prior from unpaired inputs than the latent encoder used in~\cite{chen2024hierarchical,li2024rethinking}.
(b), (c), and (h) show that combining the enhancement and transfer parts results in higher performance, verifying the necessity of each component in TPE.
Moreover, (d) and (h) demonstrate that the proposed feature space prior is more robust to feature learning and more stable for training. To demonstrate that the TPE is robust enough to benefit from different unpaired sharp images, we compare the performance of (e) sampling images guided by clustering, (f) reusing the single sharp image (taking the average across ten), and (h) sampling random sharp inputs. Owing to the memory mechanism, the TPE is able to learn beneficial textures even in random sampling. In addition, according to (g)-(i), a larger memory size $N$ contributes to the learning of more diverse textures, but reduces model efficiency. We set $N$ as 256 in our model after trade-offs.

\begin{figure}
  \begin{minipage}{0.23\textwidth}\small
    \centering
    \setlength{\tabcolsep}{0.1mm}{
    \renewcommand{\arraystretch}{1.0}
    \begin{tabular}{lcc}
    \toprule
    Method    & PSNR  & SSIM   \\
    \midrule
    \myrowcolour%
    \textit{w/o} DM &  26.46   & 0.867  \\
    Memory Bank\cite{wu2018unsupervised} &   27.89  &  0.894 \\
    \myrowcolour%
    Sparse Coding\cite{fan2020neural} & 26.66  &  0.871 \\
    Vanilla VQ\cite{van2017neural} &   27.82  & 0.895 \\
    \myrowcolour%
    DM (Ours) &  \textbf{28.13} & \textbf{0.903} \\
    \bottomrule
    \end{tabular}
    \vspace{-3mm}
    \captionof{table}{Results of different prior generation methods.}
    \label{tab:Diff}}
    \end{minipage}
  \hfill
  \begin{minipage}{0.24\textwidth}
      \centering
        \includegraphics[width=1.0\linewidth,page=6]{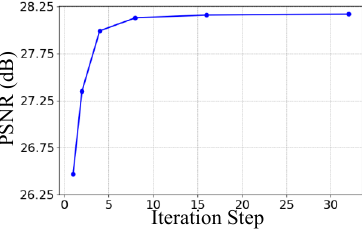}
      \vspace{-6mm}
       \caption{Ablation study of the number of iteration steps $T$.}
       \label{fig:case_ablationT}
  \end{minipage}
\end{figure}

\vspace{1mm}
\noindent\textbf{Effectiveness of DM.}
To demonstrate the necessity of DM in generating texture prior, we compare it with other methods~\cite{wu2018unsupervised,fan2020neural,van2017neural} in \cref{tab:Diff}. The DM performs favorably against the other approaches, proving the advantage of generating texture prior in unpaired restoration.
In addition, in~\cref{fig:case_ablationT}, the performance is positively correlated with the iteration steps $T$ and the gain gradually decreases when $T$ is over 8.
It shows that only a few iterations are needed to reconstruct the texture prior. We set $T$ to 8 after trade-offs.

\vspace{1mm}
\noindent\textbf{Effectiveness of FM-MSA.}
We compare (a) without filtering, (b) vanilla filtering, (c) deformable filtering~\cite{dai2017deformable}, and (d) separable filtering~\cite{su2019pixel} in \cref{tab:fil2} to show the reliability of proposed adaptive filtering within FM-MSA in~\cref{fig:overview}(c). They either cannot adaptively change the weights of different regions or are not conducive to capturing non-local blur.
Our method allows for handling more complex blurs and performs favorably against other approaches.

\begin{table}[t]\small
\centering
\begin{tabular}{l|ccccc}
\toprule
Methods     &\cellcolor{LightGray}(a)  & (b) & \cellcolor{LightGray}(c) & (d) & \cellcolor{LightGray}Ours\\
\midrule
PSNR & \cellcolor{LightGray}27.81& 27.93& \cellcolor{LightGray}28.04&28.02  & \cellcolor{LightGray}\textbf{28.13}\\
SSIM & \cellcolor{LightGray}0.895& 0.898&\cellcolor{LightGray}0.901 &0.900 &\cellcolor{LightGray}\textbf{0.903} \\
\bottomrule
\end{tabular}
\vspace{-2mm}
\caption{Results comparisons of different filtering methods.}
\label{tab:fil2}
\vspace{-2mm}
\end{table}

\section{Conclusion}
In this paper, we propose a novel diffusion model-based framework for learning texture priors (\ours).
In particular, we recover deblurred images by performing a diffusion model to generate the spatially varying texture prior.
To achieve this, we introduce the TPE to encode texture prior of the blurry input, and the TTformer layer to recover texture details using the texture prior.
Such a design fully exploits the texture knowledge in unpaired sharp images and provides inspiration for other unpaired restoration tasks. Extensive experiments demonstrate that our \ours outperforms existing SOTA methods.

\clearpage
\clearpage
\setcounter{page}{1}
\setcounter{section}{0}
\setcounter{equation}{0}
\setcounter{figure}{0}
\setcounter{table}{0}
\renewcommand{\thesection}{A\arabic{section}}
\renewcommand{\thefigure}{A\arabic{figure}}
\renewcommand{\thetable}{A\arabic{table}}
\renewcommand{\theequation}{A\arabic{equation}}
\maketitlesupplementary

In this supplementary material,~\cref{A} illustrates the detailed architecture in our TP-Diff. 
\cref{B} describes the detailed training and inference algorithms.
\cref{CC} analyses the model efficiency.
\cref{DD} describes in detail the difference between the texture prior in our method and HiDiff~\cite{chen2024hierarchical}. 
\cref{EE} provides a detailed explanation of the self-enhancement strategy mentioned in the experiments.
\cref{FF} analyses the upper bound of the performance.
\cref{G} describes the dataset used in our method.
\cref{E} analyzes the limitations. 
Finally,~\cref{F} shows more quantitative and qualitative comparison results.

\section{Architecture Details}
\label{A}

As described in Sec.~\ref{overall} of the main paper.
The deblurring network and reblurring network together form the entire cycle structure designed for removing and synthesizing blur, respectively. Within the deblurring network, to fully leverage the texture prior and enhance the model capacity, we incorporate the Texture Transfer Transformer (TTformer) at multiple scales and feed the texture prior $\hat{z}$ into them.

\begin{figure}[t]
\begin{center}
\includegraphics[width=1.0\linewidth,page=1]{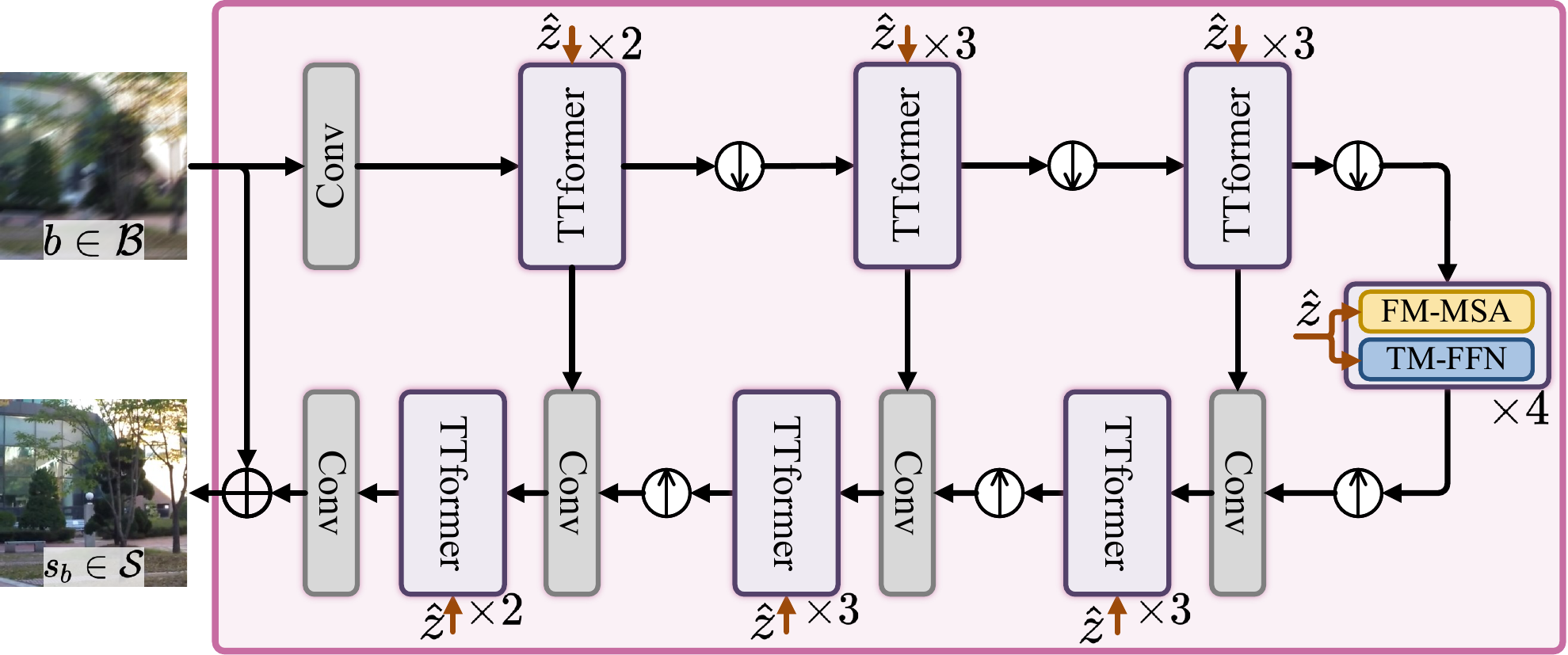}
\end{center}
\vspace{-6mm}
\caption{Network structure of deblurring network.}
\label{fig:supp-net}
\end{figure}

Specifically, we illustrate the detailed architecture of the deblurring network as shown in Fig.~\ref{fig:supp-net}. We follow the existing approach~\cite{zamir2022restormer} to learn features by stacking some TTformer layers on each scale, where the number of layers is marked.
In each TTformer layer, a filter-modulated multi-head self-attention (FM-MSA, see~\cref{fig:overview}(c) of the main paper) and a transform-modulated feed-forward network (TM-FFN, see~\cref{fig:overview}(d) of the main paper) are included. The parameters of the deblurring network are 11.8M. The reblurring network is based on the standard U-Net structure of residual blocks with a parameter size of 29.2 MB, and it is used only during training.

In addition, we use a neural network consisting of five stacked ResBlocks, denoted as $\epsilon_\theta$, to estimate the noise. The purpose of using ResBlocks as the denoising network is to ensure the same resolution of inputs and outputs while minimizing the model parameters. The parameters of the denoising network are 0.1M.

\section{Algorithm}
\label{B}

The first and second stage training algorithms for TP-Diff are shown in Alg.~\ref{alg:train_S1} and Alg.~\ref{alg:train_S2}, respectively. The inference algorithm for TP-Diff is shown in Alg.~\ref{alg:inference}.

\begin{algorithm}[t]
	\caption{ TP-Diff Training: Stage One}
	\label{alg:train_S1}
	\textbf{Input}: Texture Prior Encoder (TPE), deblurring network, reblurring network. \\
	\textbf{Output}: Trained TPE, traiend deblurring network, trained reblurring network. \\
        \vspace{-4mm}
	\begin{algorithmic}[1] 
            \FOR{$s\in\mathcal{S}$,  $b\in\mathcal{B}$ }
            \STATE    ${z}=\operatorname{TPE}(s,b). $ (paper Eqs.~(\textbf{\textcolor{blue}{1}})-(\textbf{\textcolor{blue}{4}}))
            \STATE $s_b = \operatorname{DeblurringNetwork}(b,{z})$
            \STATE $b_s = \operatorname{ReblurringNetwork}(s)$
            \STATE    ${z}=\operatorname{TPE}(s_b,b_s). $ (paper Eqs.~(\textbf{\textcolor{blue}{1}})-(\textbf{\textcolor{blue}{4}}))
            \STATE $\hat{s} = \operatorname{DeblurringNetwork}(b_s,{z})$
            \STATE $\hat{b} = \operatorname{ReblurringNetwork}(s_b)$
            \STATE Calculate $\mathcal{L}_{s1}$ loss (paper Eq.~(\textbf{\textcolor{blue}{12}})).
            \ENDFOR     
		\STATE Output the trained TPE, traiend deblurring network, trained reblurring network.
	\end{algorithmic}
\end{algorithm}

\begin{algorithm*}[t]
	\caption{ TP-Diff Training: Stage Two}
	\label{alg:train_S2}
	\textbf{Input}: Trained TPE, traiend deblurring network, trained reblurring network, denoising network, $\beta_t (t\in[1,T])$. \\
	\textbf{Output}: Trained denoising network, trained deblurring network. \\
        \vspace{-4mm}
	\begin{algorithmic}[1] 
		\STATE Init: $\alpha_t=1-\beta_t$, $\bar{\alpha}_T=\prod_{i=0}^T \alpha_i$.
            \STATE Init: The deblurring network copies the parameters of trained deblurring network. 
            \STATE Init: The reblurring network copies the parameters of trained reblurring network. 
            \STATE Init: The TPE copies the parameters of trained TPE and freezes them. 
            \FOR{$s\in\mathcal{S}$,  $b\in\mathcal{B}$ }
            \STATE    ${z}=\operatorname{TPE}(s,b). $ (paper Eqs.~(\textbf{\textcolor{blue}{1}})-(\textbf{\textcolor{blue}{4}}))
            \STATE \textbf{Diffusion Process}:
            \STATE  We sample ${z}_{T}$ by $q\left({z}_T \mid               {z}\right)=\mathcal{N}\left({z}_T; \sqrt{\bar{\alpha}_T} {z},\left(1-\bar{\alpha}_T\right) \mathbf{I}\right)$  (paper Eq.~(\textbf{\textcolor{blue}{13}})) 
            \STATE \textbf{Denoising Process}:
            \STATE $\hat{{z}}_T = {z}_{T}$
            \STATE $c=\operatorname{Conv}(b)$
            \FOR{$t=T$ to $1$ }
            \STATE $\hat{{z}}_{t-1}\!=\!\frac{1}{\sqrt{\alpha_t}}(\hat{{z}}_t-\frac{1-\alpha_t}{\sqrt{1-\bar{\alpha}_t}} \epsilon_\theta(\hat{{z}}_t, c, t))\!+\!\sqrt{1\!-\!\alpha_t} \epsilon_t$
            (paper Eq.~(\textbf{\textcolor{blue}{15}}))            
            \ENDFOR
            \STATE $\hat{{z}}=\hat{{z}}_{0}$
            \STATE $s_b = \operatorname{DeblurringNetwork}(b,\hat{{z}})$
            \STATE $b_s = \operatorname{ReblurringNetwork}(s)$
            \STATE    ${z}=\operatorname{TPE}(s_b,b_s). $ (paper Eqs.~(\textbf{\textcolor{blue}{1}})-(\textbf{\textcolor{blue}{4}}))
            \STATE \textbf{Diffusion Process}:
            \STATE  We sample ${z}_{T}$ by $q\left({z}_T \mid               {z}\right)=\mathcal{N}\left({z}_T; \sqrt{\bar{\alpha}_T} {z},\left(1-\bar{\alpha}_T\right) \mathbf{I}\right)$  (paper Eq.~(\textbf{\textcolor{blue}{13}})) 
            \STATE \textbf{Denoising Process}:
            \STATE $\hat{{z}}_T = {z}_{T}$
            \STATE $c=\operatorname{Conv}(b_s)$
            \FOR{$t=T$ to $1$ }
            \STATE $\hat{{z}}_{t-1}\!=\!\frac{1}{\sqrt{\alpha_t}}(\hat{{z}}_t-\frac{1-\alpha_t}{\sqrt{1-\bar{\alpha}_t}} \epsilon_\theta(\hat{{z}}_t, c, t))\!+\!\sqrt{1\!-\!\alpha_t} \epsilon_t$
            (paper Eq.~(\textbf{\textcolor{blue}{15}}))            
            \ENDFOR
            \STATE $\hat{{z}}=\hat{{z}}_{0}$
            \STATE $\hat{s} = \operatorname{DeblurringNetwork}(b_s,\hat{{z}})$
            \STATE $\hat{b} = \operatorname{ReblurringNetwork}(s_b)$
            \STATE Calculate $\mathcal{L}_{s2}$ loss (paper Eq.~(\textbf{\textcolor{blue}{16}})).
            \ENDFOR     
		\STATE Output the trained denoising network and trained deblurring network.
	\end{algorithmic}
\end{algorithm*}

\begin{algorithm}[t]
	\caption{ TP-Diff Inference}
	\label{alg:inference}
	\textbf{Input}: Trained denoising network, trained dehazing network, $\beta_t (t\in[1,T])$, blurry images $b\in\mathcal{B}$. \\
	\textbf{Output}: Deblurred images $S_b$. \\
        \vspace{-4mm}
	\begin{algorithmic}[1] 
		\STATE Init: $\alpha_t=1-\beta_t$, $\bar{\alpha}_T=\prod_{i=0}^T \alpha_i$.
            \STATE \textbf{Denoising Process}:
            \STATE Sample ${z}_T \sim \mathcal{N}(0,1)$
            \STATE $\hat{{z}}_T = {z}_{T}$
            \STATE $c=\operatorname{Conv}(b)$
            \FOR{$t=T$ to $1$ }
            \STATE  $\hat{{z}}_{t-1}\!=\!\frac{1}{\sqrt{\alpha_t}}(\hat{{z}}_t-\frac{1-\alpha_t}{\sqrt{1-\bar{\alpha}_t}} \epsilon_\theta(\hat{{z}}_t, c, t))\!+\!\sqrt{1\!-\!\alpha_t} \epsilon_t$
            (paper Eq.~(\textbf{\textcolor{blue}{15}}))                 
            \ENDFOR
            \STATE $\hat{{z}}=\hat{{z}}_{0}$
            \STATE $s_b = \operatorname{DeblurringNetwork}(b,\hat{{z}})$
		\STATE Output deblurred images $s_b$.
	\end{algorithmic}
\end{algorithm}

\section{Efficiency}
\label{CC}
We report the parameters and runtime compared to other state-of-the-art methods in the main paper, this section analyzes in detail the effectiveness of the core components in our methods. In particular, during inference, the parameters of our TP-Diff are 11.89M and the computational overhead is 52.7G MACs. Notably, our computational overhead is also lower than the latest method SEMGUD (TP-Diff:{52.7G} \textit{vs.} SEMGUD:63.6G).
In our TP-Diff, the diffusion model parameter used for prior reconstruction is 0.12M and the runtime is 5.2ms when inputting 256$\times$256 on 3090 GPU, where a total of 9.2G MACs is consumed for the 8 iterations. Although we use a diffusion model, it only costs a small portion of the overall model overhead, proving the efficiency of our approach.

\section{Prior Differences Compared to HiDiff~\cite{chen2024hierarchical}}
\label{DD}
It should be emphasized that our texture prior is quite different from the sharp prior in HiDiff~\cite{chen2024hierarchical}, and the reasons are as follows:
\begin{itemize}
    \item The capability of the obtained prior is different. Our texture prior in different spaces is only used to handle blurring in the corresponding region. The spatial diversity in our texture prior is reflected in that the prior with different regions is only used to handle the corresponding region. In contrast, HiDiff uses a set of out-of-order priors with a specific quantity, it cannot explicitly represent the blurring in different regions. We compare their performance in~\cref{tab:te} of the main paper, demonstrating the advantages of the generated prior in our TP-Diff.
    \item The application scenarios are different. The supervision used to generate prior in HiDiff comes from paired data and is not feasible for unpaired inputs. Benefiting from our TPE, TP-Diff can learn texture priors from unpaired data and is robust enough for different sharp inputs. Please note that it is the first attempt to introduce the diffusion model to unpaired restoration and could inspire other unpaired tasks.
    \item The structure of the denoising network used to generate the prior is different. Our TP-Diff uses CNNs to compose the denoising network, while the HiDiff uses the MLPs. In contrast, our denoising network has fewer parameters (TP-Diff: 0.12M \textit{vs.} HiDiff: 0.44M) and comparable runtimes (TP-Diff: 5.2ms \textit{vs.} HiDiff: 3.4ms) when inputting 256$\times$256 on 3090 GPU. 
\end{itemize}

\section{About Self-Enhancement Strategy in SEMGUD~\cite{chen2024unsupervised}}
\label{EE}

In~\cref{tab:main1} of the main paper, the latest SEMGUD~\cite{chen2024unsupervised} proposes a self-enhancement strategy that obtains favorable performance, this approach lacks fairness by introducing pre-trained fully supervised models to guide model training. As stated in Sec.~D of SEMGUD's supplementary, for training stability, it introduces the pre-trained NAFNet (33.69 dB PSNR on GoPro) as the extra deblurring model before estimating the prior from blurry inputs, thus bringing more performance gains. In contrast, TP-Diff is more fair by training directly from scratch using unpaired data. Therefore, we train another version of our model which is optimized with a similar strategy named TP-Diff-\textit{se} for fair comparisons. The experimental results show that we also obtain better performance when using the same strategy (TP-Diff-\textit{se}:30.16dB \textit{vs.} SEMGUD:29.06dB).

\section{About Upper Bound}
\label{FF}
It is worth emphasizing that in the first stage (\ie,~not involving the diffusion model), our model uses unpaired blurry-sharp images as input. In this case, the model performance is limited by the selection of unpaired sharp images, and the performance reaches an upper bound if fully paired blurry-sharp images are used directly as input. Theoretically, this also represents the upper bound of the second stage can be reached. If paired data inputs are used directly, the model performance reaches an upper bound (GoPro: 33.46dB/0.965, HIDE: 31.52dB/0.945). Moreover, it can also be noted from the results of HiDiff in~\cref{tab:main1} of the main paper, our method also generates a more beneficial texture prior when using fully paired inputs and yields better results.

\section{More Dataset Details}
\label{G}
We evaluate the our method on widely-used datasets: \textbf{GoPro}~\cite{nah2017deep}, \textbf{HIDE}~\cite{shen2019human}, \textbf{RealBlur}~\cite{rim2020real}, \textbf{RB2V\!\_~\!Street}~\cite{pham2023hypercut}, and \textbf{RSBlur}~\cite{rim2022realistic}. 
\textbf{GoPro}~\cite{nah2017deep} dataset includes 2,103 pairs for training and 1,111 pairs for testing. 
\textbf{HIDE}~\cite{shen2019human} dataset only includes 2,025 images pairs for testing. 
\textbf{RealBlur}~\cite{rim2020real} dataset contains two subsets: \textbf{RealBlur-R} and \textbf{RealBlur-J}. Each subset contains 980 pairs for testing. \textbf{RB2V\!\_~\!Street}~\cite{pham2023hypercut} dataset includes 9,000 pairs for training and 2,053 pairs for testing. \textbf{RSBlur}~\cite{rim2022realistic} dataset includes 8,878 pairs for training and 3,360 pairs for testing.

During training, our method requires unpaired blurry image sets $\mathcal{B}$ and sharp image sets $\mathcal{S}$.
For fair comparisons, we follow existing works~\cite{jiang2023uncertainty,chen2024unsupervised,pham2024blur2blur} to construct training data. 
Specifically, we split the training set of GoPro (containing 2,103 image pairs), RSBlur (containing 13,358 image pairs), and RB2V\!\_~\!Street (containing 11,000 image pairs) dataset into two disjoint subsets that capture different scenes with a specific ratio of 0.6:0.4. In the first subset, we select blurry images to form the blurry image set $\mathcal{B}$, while in the second subset, we choose sharp images to construct the sharp set $\mathcal{S}$. The statistics of training image sets and test image sets are reported in~\cref{tab:dataset}.

Based on this, we conduct three sets of experiments: i) Using the GoPro training set for training and the test sets for GoPro, HIDE, RealBlur-R, and RealBlur-J for testing. ii) Using the RB2V\!\_~\!Street training set for training and its test set for testing. iii) Using the RSBlur training set for training and its test set for testing.

\section{Limitation}
\label{E}

Although our texture prior can handle spatially varying blur, the resolution of the texture prior that needs to be generated increases as the input resolution increases. This means that the computational effort of the diffusion model will increase. Therefore, it is expected to make the diffusion model learn a set with a fixed number of texture priors to learn sharp features so as to avoid increasing computational costs significantly. 

In addition, a more powerful reblurring is one of the important factors in improving performance. However, the core of TP-Diff enables a powerful DM to assist the deblurring process by predicting the unknown texture prior. To realize this, we propose TPE to supervise DM training and learn to generate spatially varying texture priors. Future we will further explore the DM for reblurring performance.

\section{More Results}
\label{F}
In this section, we first provide experiments to verify the effectiveness of the diffusion model. We then analyze the sensitivity of the hyper-parameters in the loss function. Finally, we show more visualization results.

\begin{table}
    \centering
    \begin{tabular}{lccc}
    \toprule
         \multirow{2}{*}{Datasets}& \multicolumn{3}{c}{Number of data samples} \\
         \cmidrule[0.1pt](lr{0.125em}){2-4}
         \myrowcolour%
         & Train-$\mathcal{B}$  & Train-$\mathcal{S}$ & Test Pairs\\
         \midrule
         GoPro~\cite{nah2017deep}& 1,262 & 841 & 1,111\\
         \myrowcolour%
         HIDE~\cite{shen2019human}& - & - & 2,025\\
         RealBlur-R~\cite{rim2020real}&-  & - &980 \\
         \myrowcolour%
         RealBlur-J~\cite{rim2020real}& - & - & 980\\
         \midrule
         RB2V\!\_~\!Street~\cite{pham2023hypercut}& 5,400 & 3,600 & 2,053\\
         \midrule
         \myrowcolour%
         RSBlur~\cite{rim2022realistic}& 8,115 & 5,410 & 3,361\\
    \bottomrule
    \end{tabular}
    \caption{Statistics of datasets used in our method.}
    \label{tab:dataset}
\end{table}

\begin{figure}[h]
      \centering
        \includegraphics[width=0.9\linewidth,page=8]{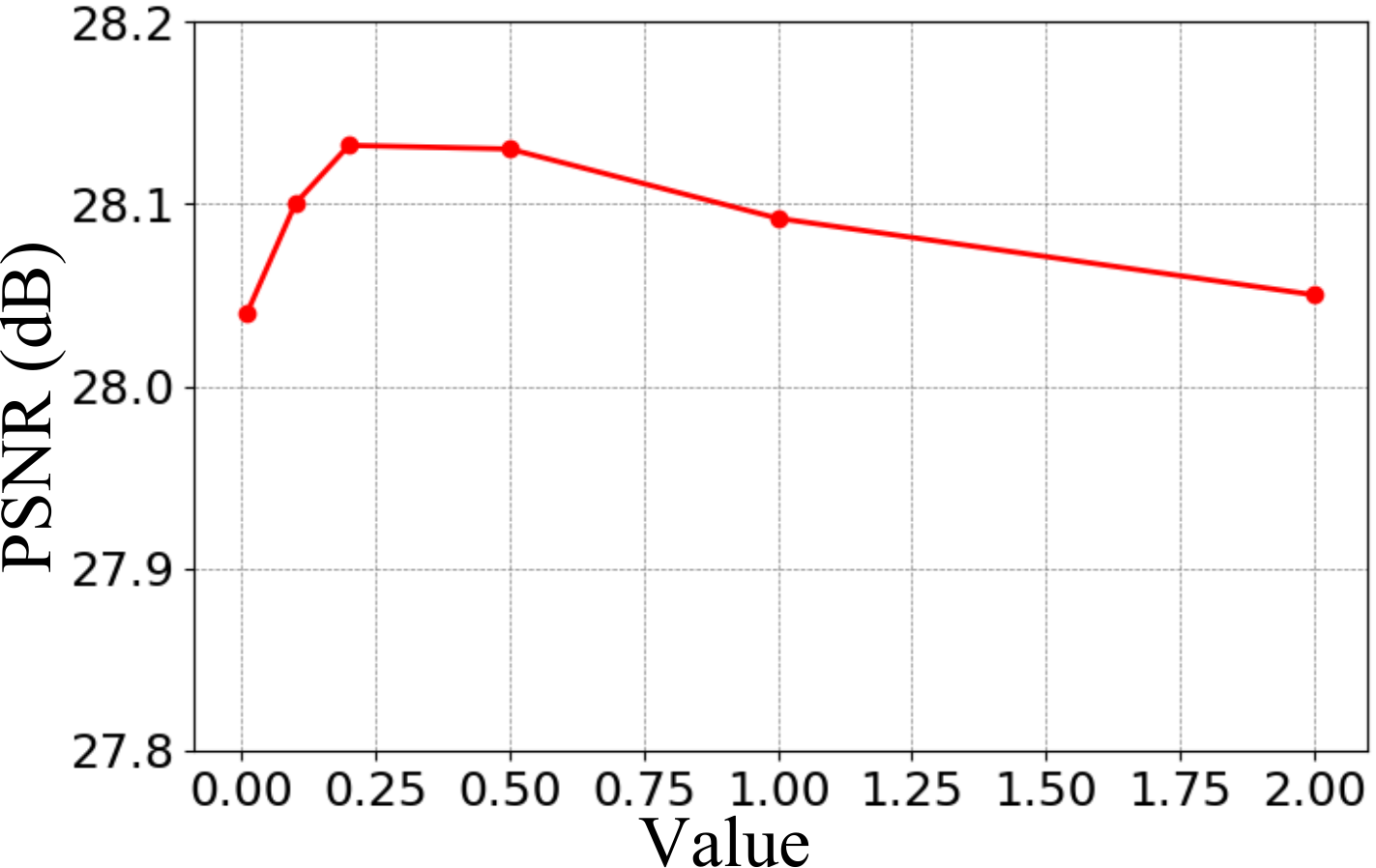}
       \caption{Sensitivity analysis of $\lambda_{Wave}$.}
       \label{fig:loss}
\end{figure}

\vspace{-2mm}
\paragraph{Effect of Hyper-parameter $\lambda_{Wave}$.}
To explore the impact of the wavelet-based adversarial loss we presented in~\cref{losswave}, we discuss the different $\lambda_{Wave}$ as shown in~\cref{fig:loss}. The experiment results show that too small $\lambda_{Wave}$ cannot effectively preserve the texture structure, while too large $\lambda_{Wave}$ affects the illumination of the image and reduces the performance. Therefore, we empirically set $\lambda_{Wave}$ to 0.2 in our model.

\begin{figure}[h]
      \centering
        \includegraphics[width=0.8\linewidth,page=7]{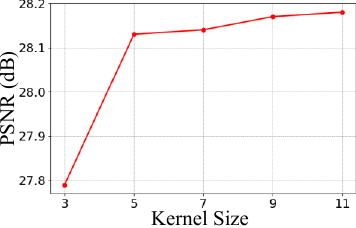}
       \caption{Effect of the number of kernel size $K$.}
       \label{fig:case_ablationK}
\end{figure}

\vspace{1mm}
\paragraph{Effect of Hyper-parameter $K$.}
To show the reliability of adaptive filtering within FM-MSA in~\cref{fig:overview}(c) of the main paper, we analyze the effect of kernel size $K$ on describing complex blurs for adaptive filtering in~\cref{fig:case_ablationK}.
The performance positively correlates with $K$. It demonstrates the powerful potential of our adaptive filtering to handle complex blurs. Although a larger $K$ will allow more pixels to be referenced, it will also increase the computational overhead. We finally set $K$ to 5.

\vspace{1mm}
\paragraph{Experiments of Cross-Validation.}
In~\cref{tab:cross-eval}, we follow~\cite{kimcontrollable,kim2024real} using RealBlur-J and RSBlur for cross-validation to verify the generalization ability. Results show that our TP-Diff is able to achieve better generalization ability compared to other unpaired training methods.  It is worth noting that it is unfair to compare the cross-validation results of our method with other generalized deblurring methods, since the unpaired inputs are already inherently more challenging than the paired inputs. In addition, the core of TP-Diff is to assist the deblurring process by introducing a diffusion model that predicts beneficial texture prior, rather than learning the blurry degradation template.

\begin{table}
    \centering
  \begin{tabular}{ l  | c  c  c}
    \hline
    Methods        &  UVCGANv2~\cite{torbunov2023uvcgan}  & UCL~\cite{wang2024ucl}  & TP-Diff \\
    \hline
    PSNR         &  24.85  &  24.56   & \textcolor{red}{25.45}\\
    \hline
    SSIM         &  0.682  &   0.701 & \textcolor{red}{0.735}\\
    \hline
  \end{tabular}
    \caption{Results of cross-validated experiments.}
    \label{tab:cross-eval}
\end{table}

\vspace{-2mm}
\paragraph{More Visual Results}

\begin{figure*}[t]
\begin{center}
\includegraphics[width=1.0\linewidth,page=2]{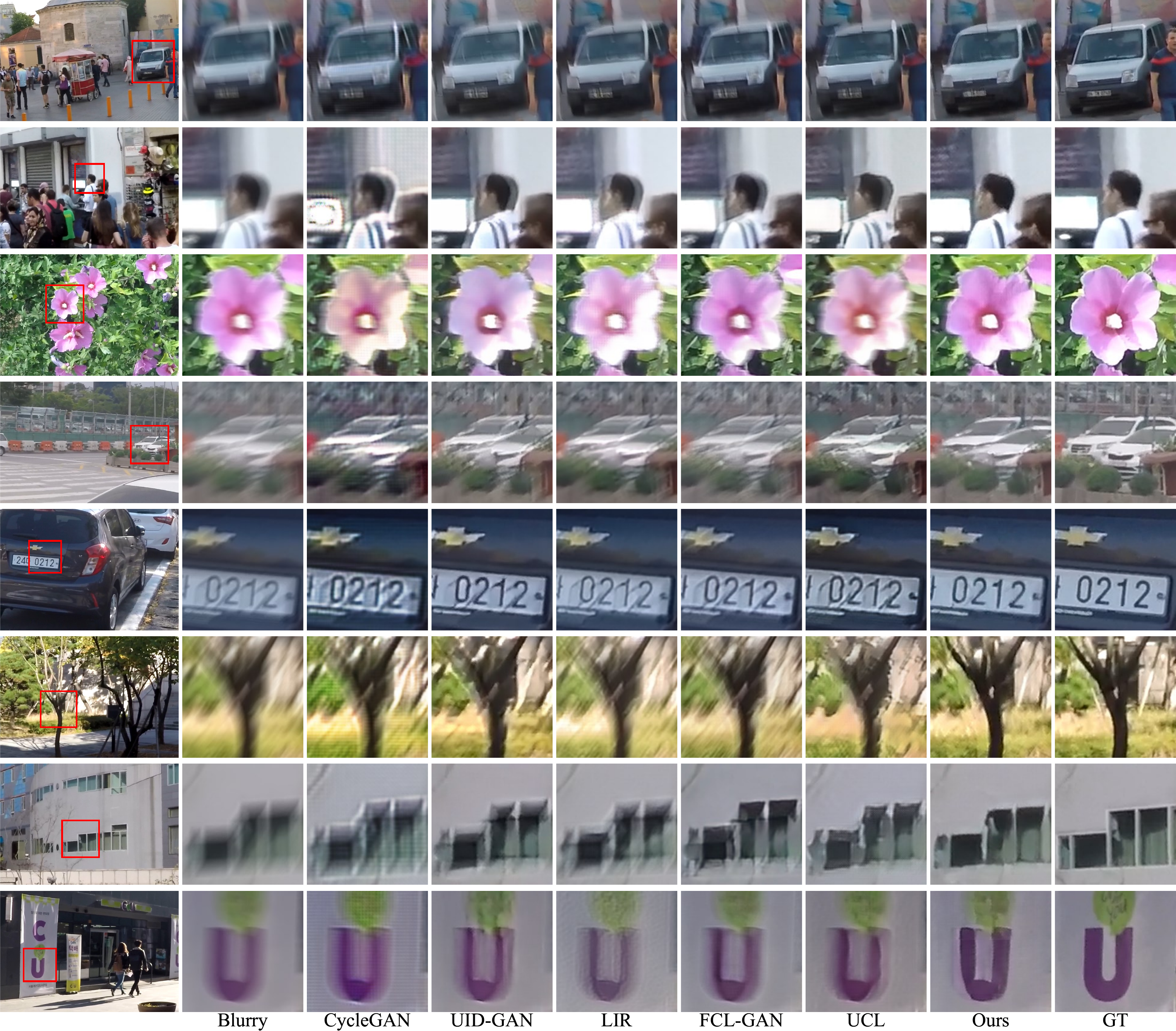}
\end{center}
\caption{Visual results on GoPro~\cite{nah2017deep} dataset. The method is shown at the bottom of each case. Zoom in to see better visualization.}
\label{fig:gopro}
\end{figure*}

\begin{figure*}[t]
\begin{center}
\includegraphics[width=1.0\linewidth,page=3]{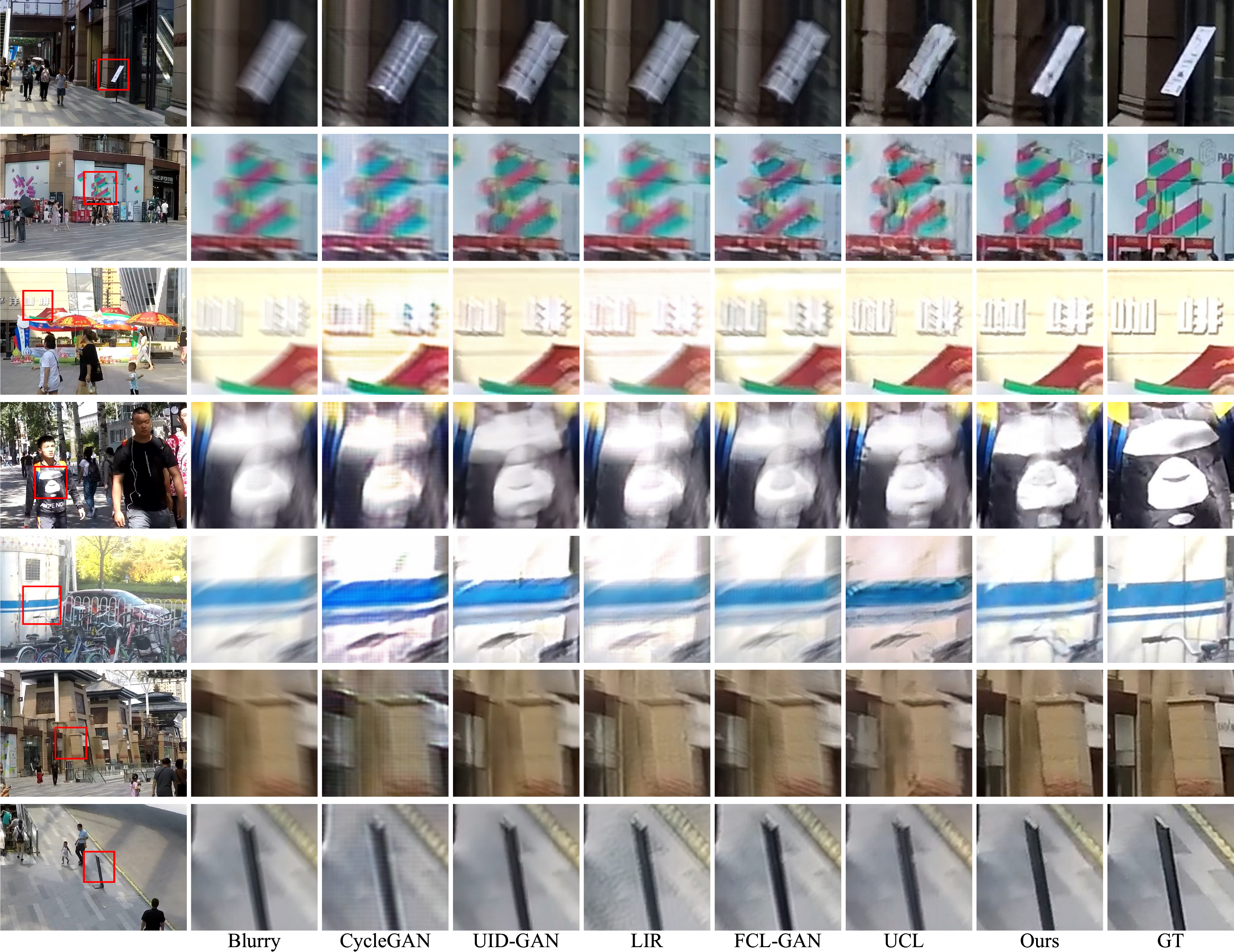}
\end{center}
\caption{Visual results on HIDE~\cite{shen2019human} dataset. The method is shown at the bottom of each case. Zoom in to see better visualization.}
\label{fig:hide}
\end{figure*}

\begin{figure*}[t]
\begin{center}
\includegraphics[width=1.0\linewidth,page=4]{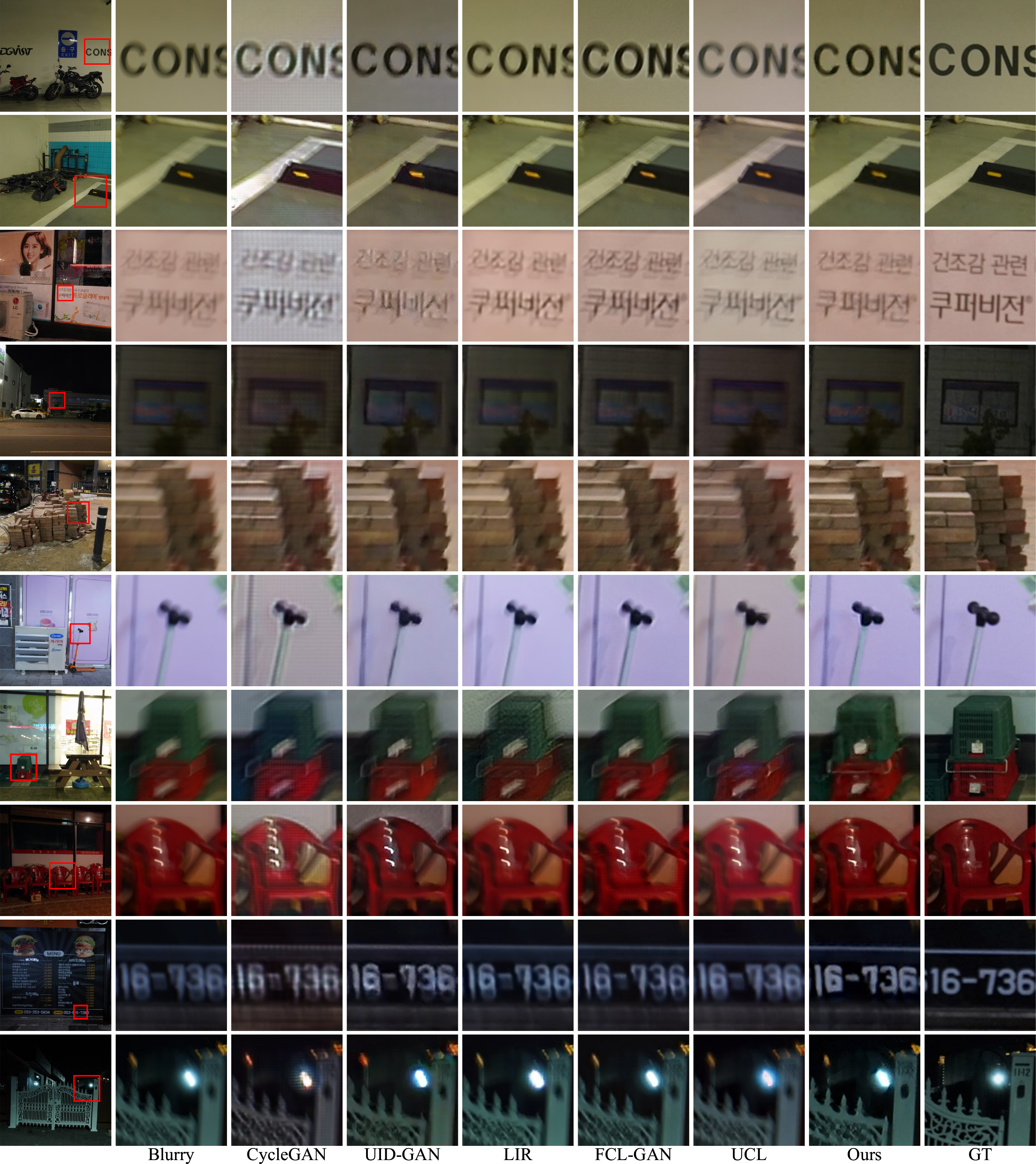}
\end{center}
\caption{Visual results on RealBlur-J~\cite{rim2020real} dataset. The method is shown at the bottom of each case. Zoom in to see better visualization.}
\label{fig:realblurj}
\end{figure*}

\begin{figure*}[t]
\begin{center}
\includegraphics[width=1.0\linewidth,page=5]{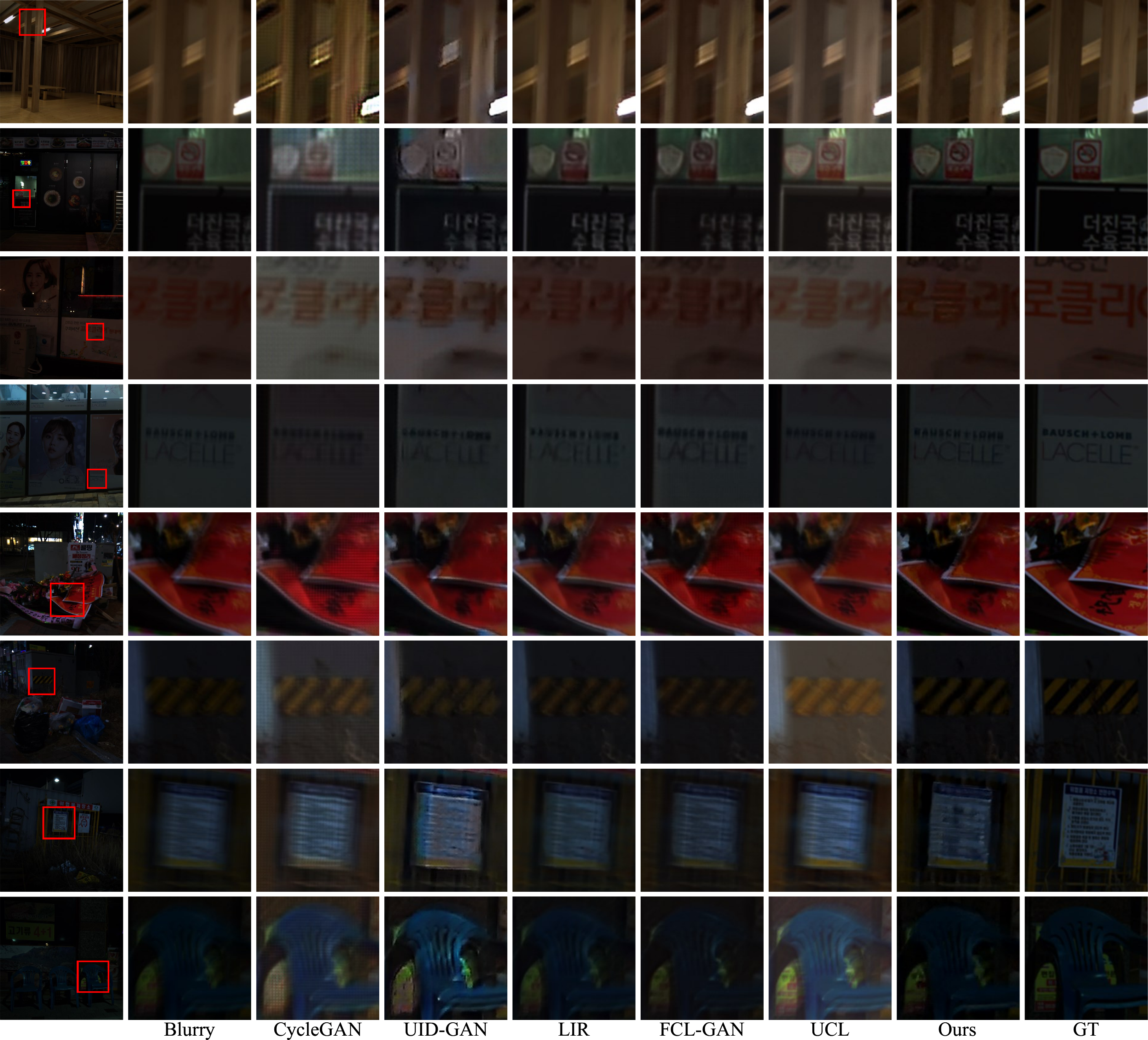}
\end{center}
\caption{Visual results on RealBlur-R~\cite{rim2020real} dataset. The method is shown at the bottom of each case. Zoom in to see better visualization.}
\label{fig:realblurr}
\end{figure*}

\begin{figure*}[t]
\begin{center}
\includegraphics[width=1.0\linewidth,page=6]{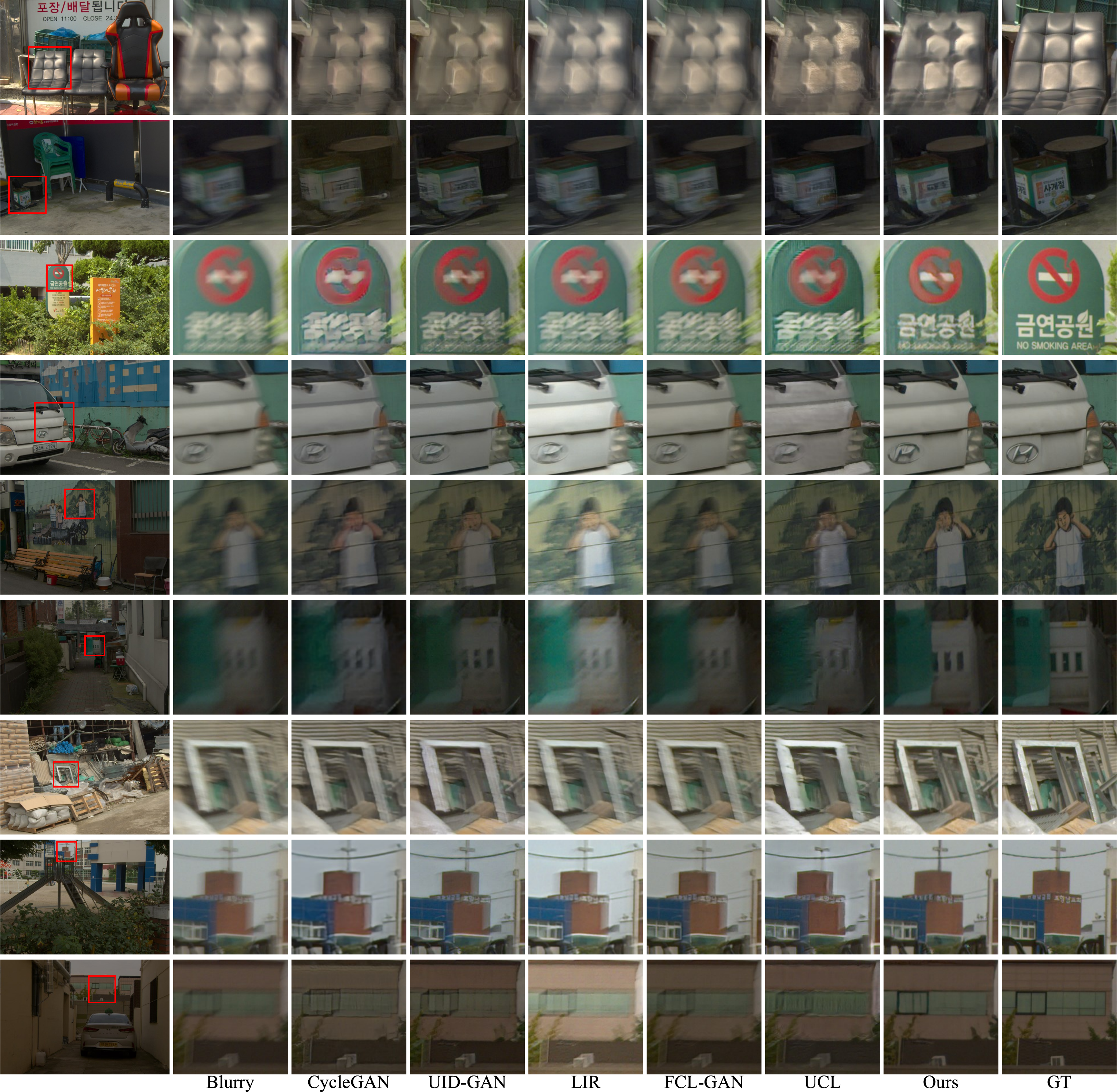}
\end{center}
\caption{Visual results on RSBlur~\cite{rim2022realistic} dataset. The method is shown at the bottom of each case. Zoom in to see better visualization.}
\label{fig:rsblur}
\end{figure*}

\begin{figure*}[t]
\begin{center}
\includegraphics[width=1.0\linewidth,page=7]{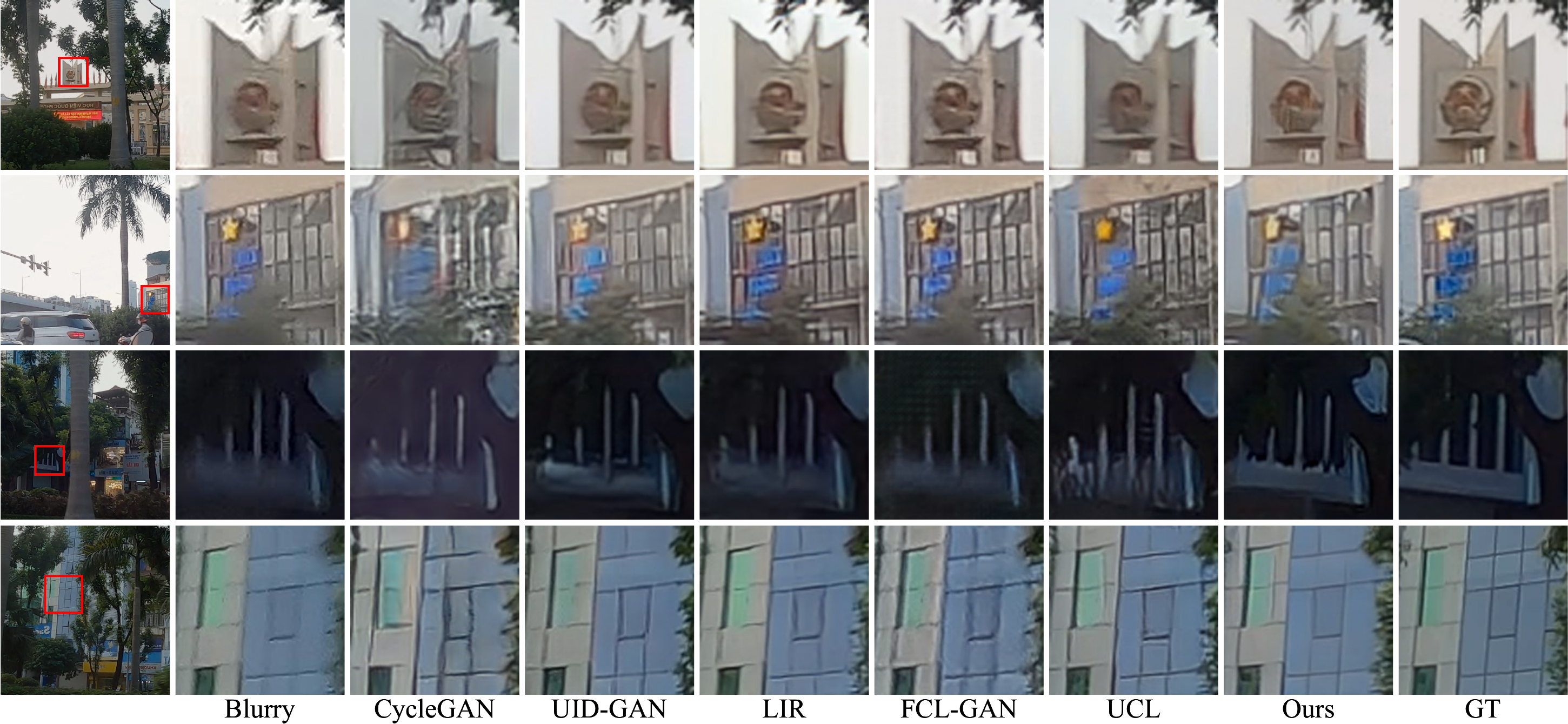}
\end{center}
\caption{Visual results on RB2V\!\_~\!Street~\cite{pham2023hypercut} dataset. The method is shown at the bottom of each case. Zoom in to see better visualization.}
\label{fig:rb2v}
\end{figure*}

To further verify the effectiveness of our method, we show more comparison results among the proposed TP-Diff and other advanced methods on six different benchmarks. The results on \textbf{GoPro}~\cite{nah2017deep}, \textbf{HIDE}~\cite{shen2019human}, \textbf{RealBlur-J}~\cite{rim2020real}, \textbf{RealBlur-R}~\cite{rim2020real}, \textbf{RSBlur}~\cite{rim2022realistic}, and \textbf{RB2V\!\_~\!Street}~\cite{pham2023hypercut} are shown in~\cref{fig:gopro},~\cref{fig:hide},~\cref{fig:realblurj},~\cref{fig:realblurr},~\cref{fig:rsblur}, and~\cref{fig:rb2v}, respectively.

\clearpage

\clearpage
\clearpage
{
    \small
    \bibliographystyle{ieeenat_fullname}
    \bibliography{main}
}

\end{document}